\documentclass[11pt]{article}

\usepackage[preprint]{acl}

\usepackage{times}
\usepackage{latexsym}

\usepackage[T1]{fontenc}

\usepackage{microtype}

\usepackage{inconsolata}

\usepackage[T1]{fontenc}    
\usepackage{hyperref}       
\usepackage{url}            
\usepackage{booktabs}       
\usepackage{amsfonts}       
\usepackage{microtype}      
\usepackage{color}
\usepackage{algorithm,algpseudocode}
\usepackage{caption}
\usepackage{mwe}
\usepackage{lipsum}

\usepackage{algorithm}
\usepackage{algpseudocode}

\usepackage{graphicx}
\usepackage{amsmath}
\usepackage{amssymb}
\usepackage{float}
\usepackage{stfloats}
\usepackage{bbm}
\usepackage{dsfont}
\usepackage{algorithm}
\usepackage{array}
\usepackage[caption=false,font=normalsize,labelfont=sf,textfont=sf]{subfig}
\usepackage{textcomp}

\usepackage{verbatim}
\usepackage{multirow}
\usepackage{graphicx}
\usepackage{pifont}
\usepackage{color}
\usepackage[hang,flushmargin]{footmisc} 
\usepackage{booktabs}
\usepackage[normalem]{ulem}

\usepackage{soul}
\usepackage{float}
\usepackage{tcolorbox}
\usepackage{tikz}

\soulregister{\cite}7
\soulregister{\ref}7
\soulregister\eqref7

\usepackage{colortbl}

\definecolor{azure(web)(azuremist)}{rgb}{0.94, 1.0, 1.0}
\definecolor{oldlace}{rgb}{0.99, 0.96, 0.9}
\definecolor{pearl}{rgb}{0.94, 0.92, 0.84}
\definecolor{seashell}{rgb}{1.0, 0.96, 0.93}
\definecolor{silver}{rgb}{0.75, 0.75, 0.75}
\definecolor{platinum}{rgb}{0.9, 0.89, 0.89}
\definecolor{almond}{rgb}{0.94, 0.87, 0.8}
\definecolor{lightskyblue}{RGB}{173, 216, 230}

\usepackage{arydshln}

\newcommand{\think}[1]{\textcolor{blue}{\texttt{<think>}} #1 \textcolor{blue}{\texttt{</think>}}}
\newcommand{\search}[1]{\textcolor{cyan}{\texttt{<search>}} #1 \textcolor{cyan}{\texttt{</search>}}}
\newcommand{\info}[1]{\textcolor{brown}{\texttt{<information>}} #1 \textcolor{brown}{\texttt{</information>}}}
\newcommand{\answer}[1]{\textcolor{purple}{\texttt{<answer>}} #1 \textcolor{purple}{\texttt{</answer>}}}

\NewDocumentCommand{\hongru}
{ mO{} }{\textcolor{blue}{\textsuperscript{\textit{Hongru}}\textsf{\textbf{\small[#1]}}}}

\title{Mitigating Context Interference for Reliable and Efficient Search Agents}

\author{
\bf Boyang Xue$^{1,6}$, 
Bin Wu$^{2}$, 
Shuofei Qiao$^{3}$,
Sheng Wang$^{4}$,
Rui Wang$^{1,6}$,
Yiming Du$^{1,6}$,\\
\bf {Hongru Wang$^{5}$}\thanks{~Co-corresponding authors.},
Jeff Z. Pan$^{5}$,
Emine Yilmaz$^{2*}$,
Kam-Fai Wong$^{1,6*}$,
Aldo Lipani$^{2}$ \\
  $^{1}$The Chinese University of Hong Kong, 
  $^{2}$University College London \\
  $^{3}$Zhejiang University,
  $^{4}$The University of Hong Kong,
  $^{5}$The University of Edinburgh \\
  $^{6}$MoE Key Laboratory of High Confidence Software Technologies \\
  {\tt \{byxue, kfwong\}@se.cuhk.edu.hk}
}

\begin{document}
\maketitle

\begin{abstract}
Recent research empowers Large Language Models (LLMs) as multi-turn search agents to iteratively retrieve and generate outputs until complex tasks are solved.
However, the contexts of multi-turn search agents are lengthy and complex.
For example, the retrieved set of documents in each turn would inevitably introduce irrelevant information that distracts LLMs, referring to \textit{context interference}, potentially hindering the reliability and efficiency of search agents.
Therefore, we conduct a systematic study on context interference in multi-turn search agents, focusing on investigating 
i) which parts of the context of search agents will contribute to the context interference,
ii) how to refine the contexts of search agents to mitigate the interference,
and iii) can incorporating context refinement into search agent training yield further improvements.
We reveal that interference primarily arises from the latest retrieved documents.
Based on the explored findings, we then introduce a distill-based context refiner to dynamically mitigate context interference for multi-turn search agents.
Finally, we validate that incorporating context refinement into RL training pipelines of search agents can significantly enhance both reliability and efficiency.
This study highlights the importance of mitigating context interference of search agents, inspiring a novel paradigm of ``refine context and then generate'' for AI agents.
\end{abstract}

\section{Introduction}
\label{sec:intro}

Large Language Models (LLMs) have 
demonstrated strong performance in 
tackling complex tasks 
using their pretrained knowledge~\citep{gpt5,deepseekai2025deepseekr1incentivizingreasoningcapability}.
Recent work has further empowered them  
to invoke search engines to retrieve {external knowledge}, essentially training them as multi-turn search agents 
that iterate on retrieval and generation until 
tasks are solved~\citep{wang2025theoryagentstoolusedecisionmakers,jia2025fastslowtoolaugmentedthinking,huang2025reinforcedinternalexternalknowledgesynergistic}.

\begin{figure}[!t]
    \centering
    \includegraphics[width=0.9\linewidth]{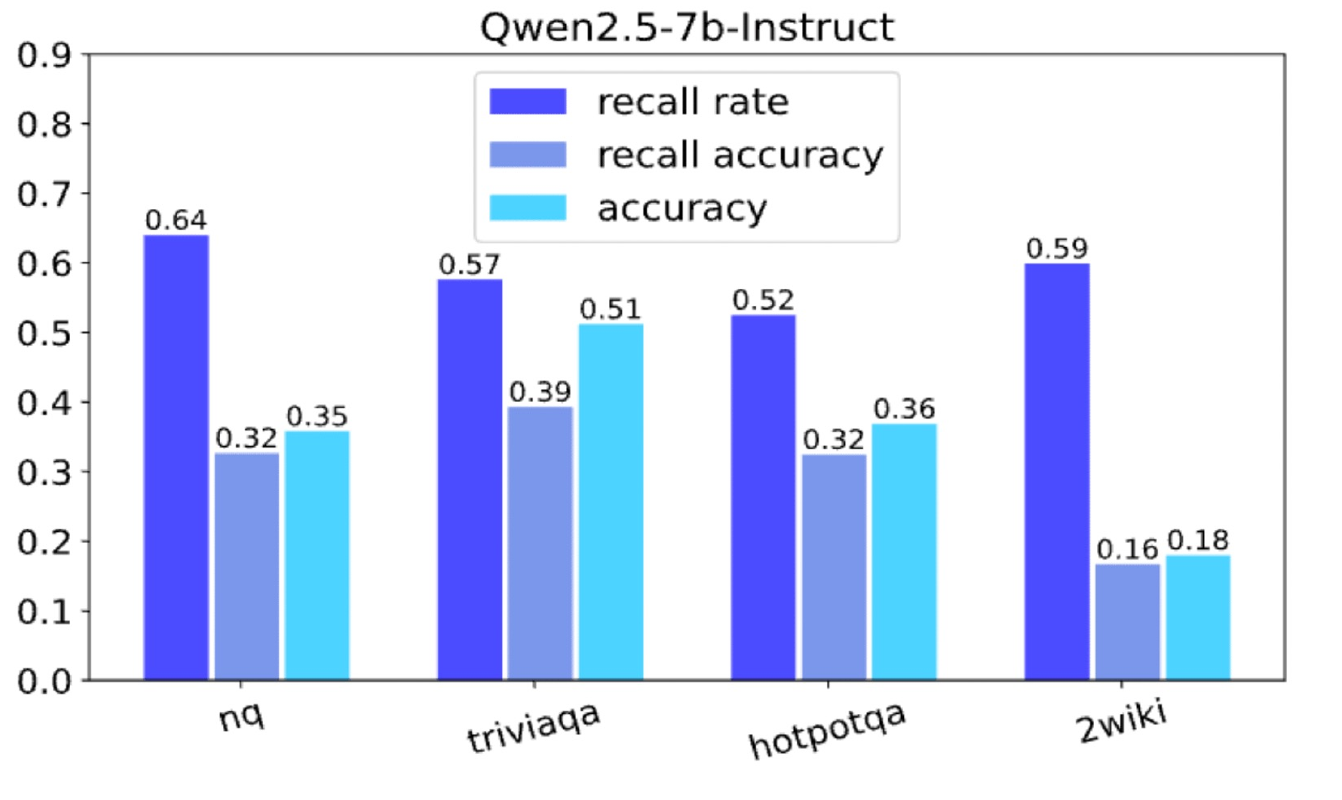}
    \vspace{-0.4em}
    \caption{Demonstration of how context interference affects the performance of LLM-based search agents. (``recall rate''=$N_\text{r}/N$ denotes the proportion of questions for which the retrieved documents contain the correct answer ($N_\text{r}$) among all questions ($N$). ``recall accuracy''=$N_\text{rc}/N$ refers to the proportion of correctly answered questions ($N_\text{rc}$) in $N_\text{r}$, relative to all questions ($N$). ``accuracy'' represents the proportion of all correctly answered questions $N_\text{c}$ out of the questions ($N$).) 
    }
    \label{fig:recall_test}
\end{figure}

However, the contexts of multi-turn search agents are lengthy and complex, encompassing the question, multi-round search queries, retrieved documents, and reasoning steps \citep{jin2025search}, which may include irrelevant information.
For instance, the retriever always returns a set of documents to ensure coverage of the search query, which also introduces noisy or irrelevant documents into the context~\citep{dong2025understand}.
This refers to \textit{context interference}, which indicates ``feeding too much irrelevant context may confuse LLMs to focus on wrong information~\citep{coleman2023incontextinterferencechatbasedlarge,haseeb2025contextengineeringmultiagentllm,gupta-etal-2024-llm,jiang2025enhancingrobustnesslargelanguage}.''
The context interference presented in each round may distract LLMs from irrelevant information and persistently degrade subsequent generation quality ~\citep{li2025singleturnsurveymultiturninteractions,laban2025llmslostmultiturnconversation}, thereby hindering the efficiency and reliability of search agents.
As in Figure~\ref{fig:recall_test}, the gaps between ``\texttt{recall rate}'' and ``\texttt{recall accuracy}'' demonstrate that search agents always retrieve the documents containing the useful information but fail to generate the correct answer, highlighting the context interference effect of accurate knowledge expression of search agents.

Prior studies to mitigate context interference have predominantly centered on dialogue systems~\citep{jacqmin-etal-2022-follow} and retrieval-augmented generation (RAG)~\citep{glass-etal-2022-re2g,nguyen2025maragmultiagentretrievalaugmentedgeneration,yu2024rankragunifyingcontextranking}, while largely overlooking techniques on multi-turn search agent settings.
Therefore, we systematically study the \textit{context interference} issue on search agents in this work with three research questions (\textbf{RQ}):

\textbf{i)}\textit{ Which parts of contexts will contribute to context interference for multi-turn search agents?}

\textbf{ii)}\textit{ How to refine contexts of search agents to mitigate such interference?}

\textbf{iii)}\textit{ Can leveraging context refinement in RL training pipelines of search agents yield further performance improvements?}

\begin{figure}[!t]
    \centering
    \includegraphics[width=0.99\linewidth]{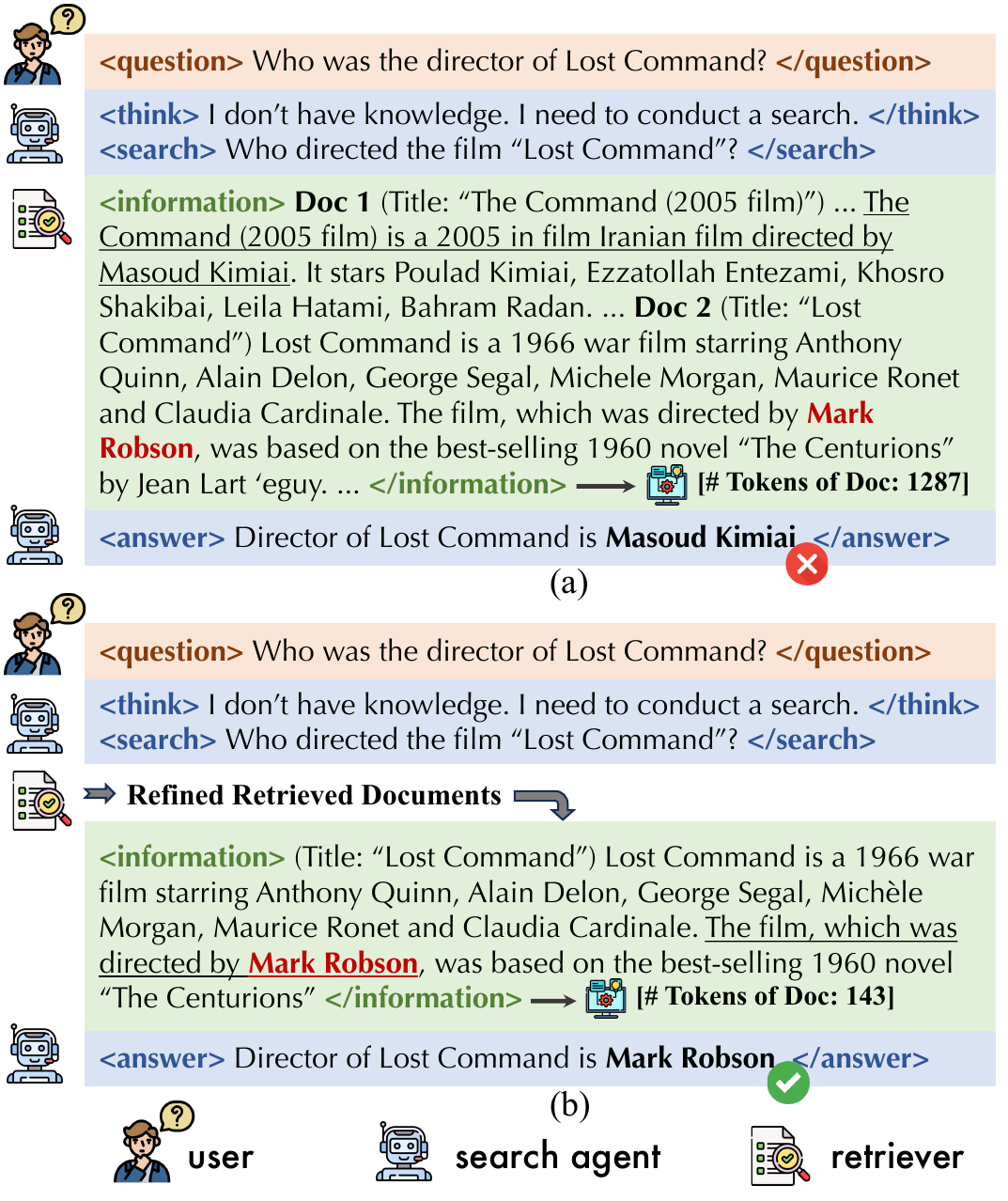}
    \caption{Examples of the search agent with (a) contextual interference in the retrieved documents and (b) refined contexts with the most critical relevant information.
    The question derives from PopQA \citep{mallen2022not}, and we employ Qwen2.5-7b-Instruct \citep{yang2024qwen2} as the foundation LLM of the search agent. We present more multi-turn QA examples of context interference mitigation in the Appendix.}
    \vspace{-0.5em}
    \label{fig:example}
\end{figure}

In light of the above questions, we investigate the context interference effect on multi-turn search agents with respect to \textbf{reliability} and \textbf{efficiency} across a series of closed-book QA benchmarks.
For \textbf{RQ i}, we compare the performances of multi-turn search agents with different inputs by masking specific parts of history contexts.
Results identify that context interference primarily derives from the latest retrieved document of search agents and slightly arises from previous search queries and documents.
For \textbf{RQ ii}, we first explore a series of context interference mitigation strategies based on the conclusion from \textbf{RQ i}.
Based on the exploration, we propose to distill a context refinement dataset comprising both retrieved documents and refined texts, which contains the most critical information in documents to the search query.
Then we train a context refiner using the dataset, which can mitigate context interference for multi-turn search agents as exemplified in Figure~\ref{fig:example} and improve both efficiency and reliability.
For \textbf{RQ iii}, we further incorporate the context refiner into the reinforcement learning (RL) training pipelines of search agents.
Experiments demonstrate that training multi-turn search agents with dynamically refined context achieves significant performance improvements regarding both the reliability and efficiency over other training baselines.

The contributions of this work are as follows:

(1) This work first investigates the context interference issue on multi-turn search agents, highlighting the necessity of context refinement to improve both efficiency and reliability of search agents, inspiring a novel paradigm of “refine context and then generate” for AI agents \footnote{We have released the codes of this work on \href{https://github.com/AmourWaltz/CRRL.git}{https://github.com/AmourWaltz/CRRL.git}.}.

(2) This work reveals that context interference primarily derives from the latest documents in multi-turn search agents, and therefore introduces a distill-based context refiner to dynamically eliminate context interference for search agents, which can be applicable to mitigate contextual interference in other search agent scenarios.

(3) This work further incorporates context refinement into RL training pipelines of search agents, which can further enhance both reliability and efficiency, providing insight into refining contexts during search agent training for future work.

\section{Preliminary of Search Agent}
\label{sec:preliminary}

To establish a theoretical foundation to analyze the context interference issue, we introduce the concepts of the search agent’s internal/external knowledge, a Markov Decision Process (MDP), and practical settings of multi-turn search agents. {{Notation definitions of this work can be found in Appendix~\ref{appendix:notation}}}.
Related works are in Appendix~\ref{append:related}.

\paragraph{Internal/External Knowledge of Search Agent}
\label{ssec:know_agent}
Previous works identify the concept of \textit{Internal/External Knowledge} for LLM agent \citep{wang2025theoryagentstoolusedecisionmakers,jia2025fastslowtoolaugmentedthinking}, where internal knowledge $\boldsymbol{\mathcal{K}}_I$ is learned from the pretrained corpus and encoded within the model parameters, and external knowledge $\boldsymbol{\mathcal{K}}_E$ is accessed through external tools (e.g., search engines).
For an LLM-based search agent $\boldsymbol{\mathcal{M}}$ with parameters $\theta$, the output $\boldsymbol{y}$ is jointly determined by its internal parametric knowledge $\boldsymbol{\mathcal{K}}_I\in\theta$, the externally retrieved documents $\boldsymbol{d}\in \boldsymbol{\mathcal{K}}_E$, and the input task $\boldsymbol{x}$ 
as \begin{math}
    \boldsymbol{y} = \boldsymbol{\mathcal{M}}_{\theta}(\boldsymbol{x}, \boldsymbol{d})
\end{math},
where $\boldsymbol{x}$ generally refers to the query and sequentially concatenated history outputs.

As $\boldsymbol{\mathcal{K}}_I$ embedded in parameters $\theta$ cannot be directly accessed, the manifestation of $\boldsymbol{\mathcal{K}}_I$ is conditioned on the context of $\boldsymbol{x}$ and $\boldsymbol{d}$.
The retriever generally returns a set of documents in $\boldsymbol{d} \in \boldsymbol{\mathcal{K}}_E$ intended to comprehensively cover the search query, but this inevitably introduces redundancy and noise.
Such extraneous content can in turn distort the utilization of both $\boldsymbol{\mathcal{K}}_I$ and $\boldsymbol{\mathcal{K}}_E$, and undermine the reliability of search agents.
In Figure~\ref{fig:recall_test}, ``recall rate'' denotes the questions calling $\boldsymbol{\mathcal{K}}_E$ containing helpful information, while ``recall accuracy'' refers to actually solved questions using $\boldsymbol{\mathcal{K}}_E$, and ``accuracy'' represents successfully answered questions through the combined use of $\boldsymbol{\mathcal{K}}_I$ and $\boldsymbol{\mathcal{K}}_E$.
The gaps between ``recall rate'' and ``recall accuracy'' / ``accuracy'' exemplify the negative impact of context interference.
Moreover, distractions from irrelevant information will incur extra retrievals and inference costs, degrading the efficiency of search agents.

\paragraph{Markov Decision Process of Agent}
\label{ssec:mdp_agent}
The process of an LLM agent $\boldsymbol{\mathcal{M}}_{\theta}$ with $\boldsymbol{\mathcal{K}}_I\in\theta$ interacting with an environment $\boldsymbol{E}$ (referred to $\boldsymbol{\mathcal{K}}_E$) to complete a task can be regarded as a Markov Decision Process: $\left ( \boldsymbol{\mathcal{S}}, \boldsymbol{\mathcal{A}}, \boldsymbol{\mathcal{T}}, \boldsymbol{\mathcal{O}}, \boldsymbol{\mathcal{R}} \right )$.
Initially, a specific task $\boldsymbol{x}$ is provided as the initial environmental state.
Assuming the interaction proceeds in $N$ rounds and in round $n$ ($n<N$), the LLM agent receives observation $\boldsymbol{o}_{n}\in\boldsymbol{\mathcal{O}}$ and takes action $\boldsymbol{a}_n\in\boldsymbol{\mathcal{A}}$.
The state $\boldsymbol{s}_n$ at round $n$ is the history context of all preceding concatenated sequences $\boldsymbol{s}_n=(\boldsymbol{x}, \boldsymbol{a}_0, \boldsymbol{o}_{1}, \boldsymbol{a}_1, \dots, \boldsymbol{a}_{n-1}, \boldsymbol{o}_{n})\in\boldsymbol{\mathcal{S}}$.
$\boldsymbol{\mathcal{M}}_{\theta}$ is responsible for deciding $\boldsymbol{a}_n$ based on $\boldsymbol{s}_n$: 
\begin{math}
    \boldsymbol{a}_n\sim \boldsymbol{\mathcal{M}}_{\theta}(\cdot|\boldsymbol{s}_n)
\end{math}, and a retriever $\boldsymbol{\mathcal{E}}$ interacts with $\boldsymbol{E}$ to determines the state transition $\boldsymbol{\mathcal{T}}$.
Upon task completion after $N$ rounds, the trajectory is characterized as $\boldsymbol{\tau}=(\boldsymbol{x}, \boldsymbol{s}_0, \boldsymbol{a}_0, \boldsymbol{o}_{1}, \dots, \boldsymbol{a}_{N-1}, \boldsymbol{o}_{N}, \boldsymbol{a}_N)$ where the final response $\boldsymbol{y}$ is parsed from $\boldsymbol{a}_N$.
A final reward $r$ is provided by the reward model $\boldsymbol{\mathcal{R}}$ by comparing $\boldsymbol{y}$ and the reference $\boldsymbol{\hat y}$ as
\begin{math}
    r= \boldsymbol{\mathcal{R}}(\boldsymbol{y}, \boldsymbol{\hat y})
\end{math}.

\paragraph{Multi-turn Search Agent}
\label{ssec:mt_agent}
For multi-turn search agents, the $i$-th observation $\boldsymbol{o}_{i}$ is generally a list of retrieved Top-$K$ documents $\boldsymbol{d}_i=[\boldsymbol{d}_{i,1}, \boldsymbol{d}_{i,2}, \dots, \boldsymbol{d}_{i,K}]$ returned by the retriever $\boldsymbol{\mathcal{E}}(\boldsymbol{q}_{i-1}|\boldsymbol{K}_b)$ where $\boldsymbol{q}_{i-1}$ is the search query generated in the previous round and $\boldsymbol{K}_b$ is the external knowledge base as $\boldsymbol{\mathcal{K}}_E$.
The action $\boldsymbol{a}_{i}$ includes a thinking step $\boldsymbol{t}_{i}$ and a search query $\boldsymbol{q}_{i}$.
Following~\citet{jin2025search}, LLMs are instructed to encapsulate their search queries, retrieved documents, and final answer between specially designated tokens respectively.
Both $\boldsymbol{q}_i$ and $\boldsymbol{d}_i$ are appended to the context in each turn.
When generating $\boldsymbol{t}_i$, all the preceding sequences in $\boldsymbol{s}_{i}$ are fed into $\boldsymbol{\mathcal{M}}_{\theta}$ and contribute to $\boldsymbol{a}_i$.
However, not all previous documents and search queries are pertinent to the current thinking step, and the inclusion of irrelevant content can introduce context interference, impairing the LLMs’ efficiency and reliability to accurately express knowledge.

\section{Context Interference in Search Agent}
\label{sec:interference}


\begin{table*}[ht]
    \centering
    \footnotesize
    {
    \begin{tabular}{lcccccccccccc}
        \toprule
        \multirow{2}{*}{\textbf{Methods}} & \multicolumn{3}{c}{\textbf{Single-Hop QA}} & \multicolumn{4}{c}{\textbf{Multi-Hop QA}} & \multirow{2}{*}{\textit{\textbf{Avg.}}} \\
        \cmidrule(r){2-4}\cmidrule(r){5-8}
         & \textbf{NQ} & \textbf{TriviaQA} & \textbf{PopQA} & \textbf{HotpotQA} & \textbf{2wiki} & \textbf{Musique} & \textbf{Bamboogle} & \\
        \hline
        \hline
        \rowcolor{seashell}
        \multicolumn{9}{c}{\textbf{{Qwen2.5-7b-Instruct}}} \\
        \hline
        \bf Direct & 17.7~/~0.0 & 44.2~/~0.0 & 15.5~/~0.0 & 17.9~/~0.0 & \textbf{24.0}~/~0.0 & 3.9~/~0.0 & 8.0~/~0.0 & 18.7~/~0.0 \\
        \bf CoT & 17.7~/~0.0 & 47.0~/~0.0 & 13.4~/~0.0 & 21.0~/~0.0 & 23.6~/~0.0 & 4.7~/~0.0 & 30.4~/~0.0 & 22.5~/~0.0 \\
        \hdashline
        \bf IRCoT & 30.6~/~2.0 & 51.2~/~1.8 & 31.7~/~2.1 & 24.6~/~3.0 & 18.0~/~3.4 & 9.8~/~3.1 & 28.8~/~2.5 & 27.5~/~2.6 \\
        \bf IRCoT-$o$ & 29.8~/~\textbf{1.8} & 51.3~/~\textbf{1.7} & 32.7~/~\textbf{1.7} & 25.3~/~2.7 & 22.7~/~3.1 & 11.2~/~2.9 & {34.4}~/~1.9 & 29.6~/~\textbf{2.3} \\
        \bf IRCoT-$oq$ & {32.8}~/~2.0 & \textbf{51.7}~/~2.0 & \textbf{33.0}~/~2.0 & \textbf{26.0}~/~3.0 & {23.6}~/~\textbf{3.1} & \textbf{11.4}~/~3.1 & {34.4}~/~\textbf{1.7} & \textbf{30.4}~/~2.4 \\
        \bf IRCoT-$oqp$ & \textbf{32.9}~/~2.6 & 51.5~/~2.7 & 32.7~/~2.8 & 25.4~/~\textbf{2.6} & 22.8~/~3.1 & 10.6~/~\textbf{2.8} & \textbf{36.0}~/~2.5 & 30.3~/~2.7 \\
        \hline
        \rowcolor{seashell}
        \multicolumn{9}{c}{\textbf{{Qwen2.5-3b-Instruct}}} \\
        \hline
        \bf Direct & 9.7~/~0.0 & 25.7~/~0.0 & 7.7~/~0.0 & 13.5~/~0.0 & 17.1~/~0.0 & 1.7~/~0.0 & 3.2~/~0.0 & 11.2~/~0.0 \\
        \bf CoT & 12.4~/~0.0 & 35.4~/~0.0 & 9.4~/~0.0 & 15.5~/~0.0 & 16.4~/~0.0 & 2.6~/~0.0 & \textbf{20.0}~/~0.0 & 16.0~/~0.0 \\
        \hdashline
        \bf IRCoT & 21.6~/~\textbf{1.2} & 45.2~/~1.1 & 29.6~/~1.1 & 24.1~/~1.6 & 23.5~/~1.9 & 6.8~/~1.6 & 19.2~/~1.4 & 24.3~/~1.4 \\
        \bf IRCoT-$o$ & 22.4~/~1.2 & 47.0~/~\textbf{1.1} & \textbf{31.8}~/~\textbf{1.1} & 24.7~/~\textbf{1.5} & 21.3~/~\textbf{1.7} & \textbf{8.0}~/~1.5 & 19.4~/~\textbf{1.2} & 24.9~/~\textbf{1.3} \\
        \bf IRCoT-$oq$ & 23.5~/~1.3 & 47.2~/~1.2 & 29.9~/~1.4 & \textbf{24.9}~/~1.6 & \textbf{24.2}~/~1.7 & 7.0~/~1.6 & {19.4}~/~1.2 & \textbf{25.2}~/~1.4 \\
        \bf IRCoT-$oqp$ & \textbf{24.6}~/~1.5 & \textbf{47.7}~/~1.3 & 30.5~/~1.5 & 24.5~/~1.8 & 22.5~/~1.7 & 5.1~/~\textbf{1.5} & 19.0~/~1.2 & 24.8~/~1.5 \\
        \bottomrule
    \end{tabular}}
    \caption{Performance (EM/ART) of different inference methods for search agents across QA test sets, measured by Exact Match (EM, for reliability) and Average Retrieval Times (ART, for efficiency).  
    }
    \label{table:main}
\end{table*}
In this section, 
we first detail the evaluation setting in Sec.~\ref{ssec:settings}. 
Then we analyze context interference effects in different parts of contexts of search agents in Sec.~\ref{ssec:components}, and finally propose the context refiner to mitigate interference in Sec~\ref{ssec:refiner}.

\subsection{Evaluation Settings}
\label{ssec:settings}

\paragraph{Dataset}
Various closed-book QA datasets are employed to evaluate the performance of search agents, 
which necessitate extra retrieval to address, encompassing both single- and multi-hop scenarios.
\textbf{Single-hop QA} includes: \textit{Natural Questions (NQ)} \citep{kwiatkowski2019natural}, \textit{TriviaQA} \citep{joshi2017triviaqa}, and \textit{PopQA} \citep{mallen2022not}.
\textbf{Multi-hop QA} includes: \textit{HotpotQA} \citep{yang2018hotpotqa}, \textit{2WikiMultiHopQA (2Wiki)} \citep{ho2020constructing}, \textit{MuSiQue} \citep{trivedi2022musique}, and \textit{Bamboogle} \citep{press2022measuring}.
Dataset details are presented in Appendix~\ref{appendix:dataset}.

\paragraph{Search Agent}
We employ two foundation LLMs $\boldsymbol{\mathcal{M}}$ for search agents: \textbf{Qwen-2.5-7b-Instruct} and \textbf{Qwen-2.5-3b-Instruct} \citep{yang2024qwen2}.
For retrieval, we leverage E5 \citep{wang2022text} as the retriever $\boldsymbol{\mathcal{E}}$ and 2018 Wikipedia dump \citep{karpukhin2020dense} as the knowledge base $\boldsymbol{K}_b$ respectively.
The number of retrieved passages $K$ is set to 3 across all retrieval-based methods.

\paragraph{Evaluation Metrics}
We employ several metrics to evaluate both the \textbf{reliability} and \textbf{efficiency} of search agents.
For reliability, we assess the correctness of the generated answer $\boldsymbol{y}$ with the reference $\boldsymbol{\hat y}$ using \textbf{Exact Match (EM)} \citep{song2025r1searcherincentivizingsearchcapability,jin2025search}, a standard string-matching metric that checks whether $\boldsymbol y\equiv \boldsymbol {\hat y}$, which is a percentage that represents the proportion of correctly answered questions out of all questions.
Efficiency is evaluated regarding different aspects.
Since the time cost of search agents is primarily determined by the number of retrieval operations, the \textbf{average retrieval times (ART)}, which denotes the average number of retrievals required per question, is employed to intuitively evaluate the efficiency of the generation processes.
In addition, the \textbf{average context length (Len.)} in multi-turn generations and the \textbf{average inference time per question (AIT)} (/seconds) are also employed to assist efficiency assessments for search agents.


\subsection{Context Interference in Different Parts of Context of Search Agents}
\label{ssec:components}

\paragraph{Background}
To figure out context interference effects of different parts in the contexts of search agents (as \textbf{RQ i}), we compare the performance of multi-turn search agents with different input contexts by masking specific segments (actions and observations in preceding rounds) of the history state.
The history state $\boldsymbol{s}_i$ includes input question $\boldsymbol{x}$ and a series of actions $\boldsymbol{a}_{0:i-1}$ and observations $\boldsymbol{o}_{1:i}$.
Specifically, $\boldsymbol{x}$ representing the initial state is fixed.
$\boldsymbol{p}_i$ in $\boldsymbol{a}_i$ generally involves summarizing and reasoning from $\boldsymbol{o}_i$.
$\boldsymbol{q}_i$ is to interact with $\boldsymbol{K}_b$ to get retrieved documents in $\boldsymbol{o}_{i+1}$.
We present several inference methods for search agents as follows.

\paragraph{Inference Methods}
We employ direct inference (\textbf{Direct}) and Chain-of-Thought (\textbf{CoT}) reasoning \citep{wei2022chain} as two retrieval-free baselines, which represent LLMs' internal knowledge $\boldsymbol{\mathcal{K}}_I$ to answer questions.
The prompt templates are in Appendix \ref{append:prompt}.
For retrieval-based settings of search agents, we employ Information Retrieval with CoT (\textbf{IRCoT}) \citep{trivedi2022interleaving}, which enables LLMs to actively call the retriever for questions beyond their knowledge scope after thinking.


To understand the effect of different parts of context, several variants based on \textbf{IRCoT} are developed.
Since search agents typically decompose a complex question into a set of sub-questions \citep{jin2025search,song2025r1searcherincentivizingsearchcapability}, the generated search queries and retrieved documents in a history state are generally mutually independent, exhibiting rarely sequential dependencies.
Therefore, we can specifically mask different parts in the states as follows.
1) To investigate interference in previous documents, \textbf{IRCoT-$o$ (\textbf{w/o} $\boldsymbol{o}_{:-1}$)} only incorporates the latest observation $\boldsymbol{o}_i$ of retrieved documents in context $[\boldsymbol{x}, \boldsymbol{p}_0, \boldsymbol{q}_1, \boldsymbol{p}_1, \boldsymbol{q}_1, \dots, \boldsymbol{q}_{i-1}, \boldsymbol{o}_{i}]$ when generating $\boldsymbol{a}_i$.
2) For interference in search queries, \textbf{IRCoT-$oq$ (w/o $\boldsymbol{o}_{:-1},\boldsymbol{q}_{:-1}$)} with context $[\boldsymbol{x}, \boldsymbol{p}_0, \boldsymbol{p}_1, \dots, \boldsymbol{p}_{i-1}, \boldsymbol{q}_{i-1}, \boldsymbol{o}_{i}]$ is also employed.
3) For previous thinking steps, we utilize \textbf{IRCoT-${oqp}$ (w/o $\boldsymbol{o}_{:-1},\boldsymbol{q}_{:-1},\boldsymbol{p}_{:-1}$)} relies exclusively on the latest thinking, search query, and observation as context $[\boldsymbol{x}, \boldsymbol{p}_{i-1}, \boldsymbol{q}_{i-1}, \boldsymbol{o}_{i}]$ when generating $\boldsymbol{a}_i$. 

\paragraph{Analysis and Findings}
As presented in Table~\ref{table:main}, benefiting from $\boldsymbol{K}_b$, all retrieval-based methods outperform retrieval-free baselines.
Moreover, \textbf{IRCoT-$o$} outperforms \textbf{IRCoT} in both reliability and efficiency, suggesting that previous retrieved documents before round $i$ contain context interference for generating $\boldsymbol{a}_i$.
\textbf{IRCoT-${oq}$} marginally outperforms \textbf{IRCoT-$o$}, indicating that previous search queries $\boldsymbol{q}_{:-1}$ also carry slight interference, although removing them may incur little extra retrieval.
\textbf{IRCoT-$oqp$} exhibits a slight drop in reliability and a notable decrease in efficiency compared with others, implying that previous thinking steps store key information for future steps.
LLMs may repeat previous search after masking $\boldsymbol{p}_{:-1}$ and thus add retrieval costs.
Consequently, \textbf{context interference when generating $\boldsymbol{a}_i$ may arise from both previous search queries and documents}. 

\begin{figure*}[!t]
    \centering
    \includegraphics[width=0.86\linewidth]{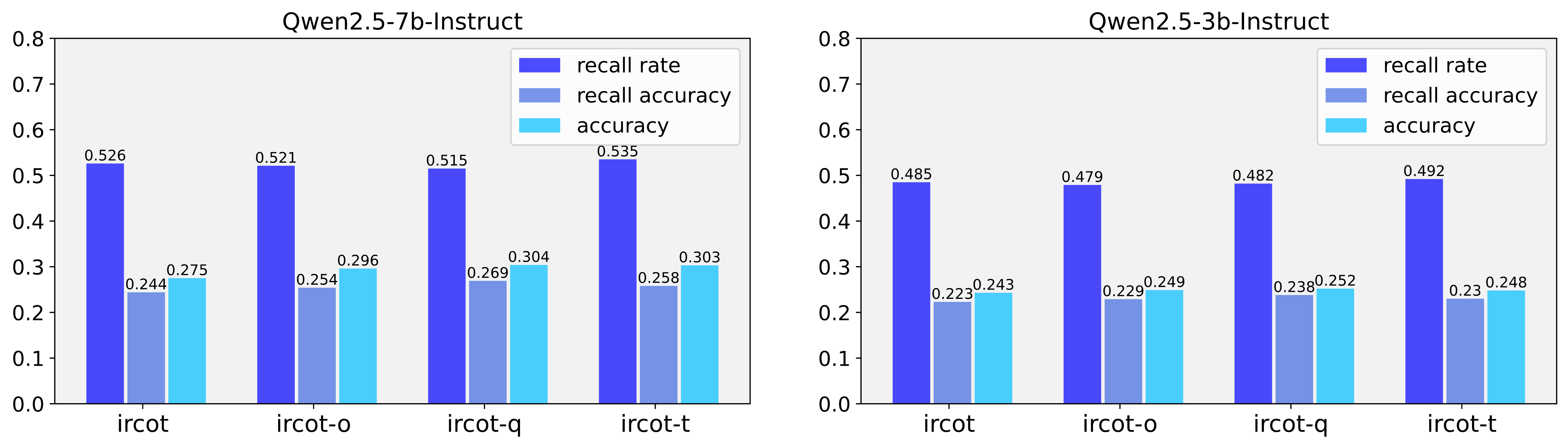}
    \caption{Demonstrations of context interference effects on four IRCoT variants of search agents. Results are averaged on all QA test sets; metrics follow the definitions in Figure~\ref{fig:recall_test}.}
    \label{fig:recall_inter}
\end{figure*}

However, as presented in Figure~\ref{fig:recall_inter}, although the above \textbf{IRCoT} variants gain slight improvements in accuracy, the gap between ``recall rate'' and ``recall accuracy'' remains considerable, indicating context interference in previous rounds is not the dominant factor.
Since the latest thinking $\boldsymbol{p}_{i-1}$ and search query $\boldsymbol{q}_{i-1}$ mostly encapsulate summaries and reasoning of previous context, it can be inferred that \textbf{the latest observation $\boldsymbol{o}_i$ is subject to the primary cause of context interference}.
To mitigate the interference, further context refinement methods for the latest observed documents are required on multi-turn search agents. 

\begin{figure*}[!ht]
    \centering
    \includegraphics[width=0.94\linewidth]{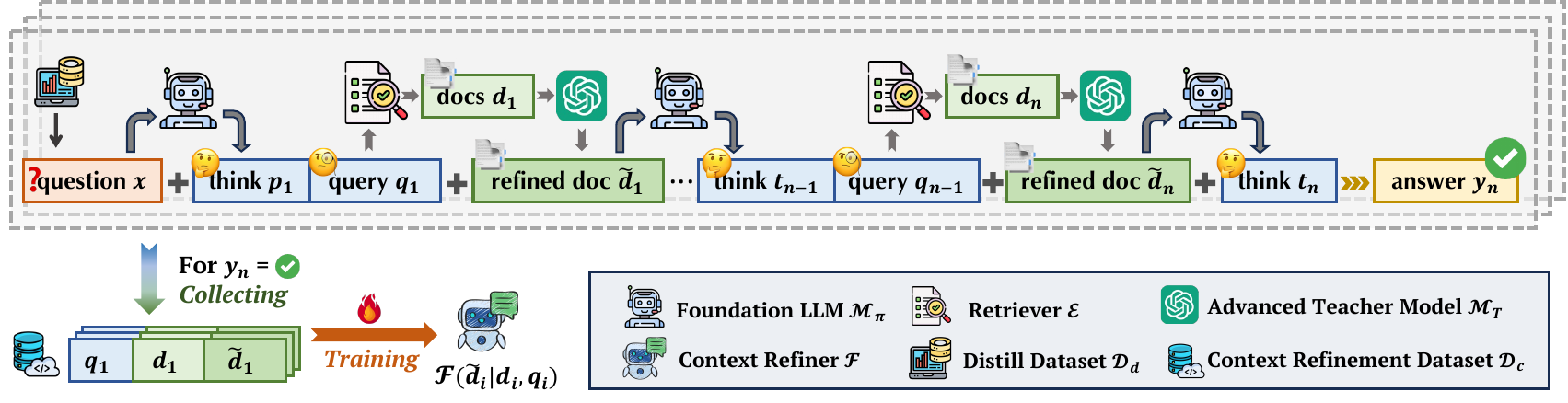}
    \caption{
    Training pipeline for the distill-based context refiner. 
    }
    \label{fig:refiner}
\end{figure*}

\subsection{Context Refiner for Search Agent}
\label{ssec:refiner}

\begin{table*}[ht]
    \centering
    \footnotesize
    {
    \begin{tabular}{lcccccccccccc}
        \toprule
        \multirow{2}{*}{\textbf{Methods}} & \multicolumn{3}{c}{\textbf{Single-Hop QA}} & \multicolumn{4}{c}{\textbf{Multi-Hop QA}} & \multirow{2}{*}{\textit{\textbf{Avg.}}} \\
        \cmidrule(r){2-4}\cmidrule(r){5-8}
         & \textbf{NQ} & \textbf{TriviaQA} & \textbf{PopQA} & \textbf{HotpotQA} & \textbf{2wiki} & \textbf{Musique} & \textbf{Bamboogle} & \\
        \hline
        \hline
        \rowcolor{seashell}
        \multicolumn{9}{c}{\textbf{{Qwen2.5-7b-Instruct}}} \\
        \hline
        \bf IRCoT & 30.6~/~2.0 & 51.2~/~1.8 & 31.7~/~2.1 & 24.6~/~3.0 & 18.0~/~3.4 & \textbf{9.8}~/~3.1 & 28.8~/~2.5 & 27.5~/~2.6 \\
        \hdashline
        \bf GPT-Compress & 32.5~/~\textbf{0.9} & 55.5~/~1.0 & 34.8~/~1.1 & 29.8~/~1.6 & 25.0~/~1.5 & 9.6~/~1.3 & 26.4~/~1.1 & 30.5~/~\textbf{1.2} \\
        \bf GPT-Refine & \textbf{34.6}~/~1.1 & \textbf{57.5}~/~\textbf{0.9} & \textbf{37.0}~/~1.0 & \textbf{33.0}~/~\textbf{1.3} & \textbf{27.6}~/~1.8 & 9.3~/~1.4 & \textbf{33.6}~/~1.0 & \textbf{33.2}~/~1.2 \\
        \bf Self-Refine & 32.5~/~1.5 & 53.3~/~0.9 & 33.0~/~\textbf{1.0} & 27.0~/~1.4 & 21.6~/~\textbf{1.5} & 8.4~/~\textbf{1.2} & 28.8~/~\textbf{1.0} & 29.2~/~1.2 \\
        \hdashline
        \bf Context Refiner & 34.0~/~1.2 & 56.9~/~0.9 & 36.4~/~1.0 & 32.0~/~1.4 & 27.2~/~1.6 & 8.6~/~1.2 & 30.4~/~1.0 & 32.2~/~1.2 \\
        \hline
        \rowcolor{seashell}
        \multicolumn{9}{c}{\textbf{{Qwen2.5-3b-Instruct}}} \\
        \hline
        \bf IRCoT & 21.6~/~1.2 & 45.2~/~1.1 & 29.6~/~1.1 & 24.1~/~1.6 & \textbf{23.5}~/~1.9 & 6.8~/~1.6 & 19.2~/~1.4 & 24.3~/~1.4 \\
        \hdashline
        \bf GPT-Compress & 31.3~/~\textbf{0.9} & 51.5~/~0.9 & 29.9~/~1.0 & 24.1~/~1.1 & 22.7~/~\textbf{1.2} & 7.2~/~1.0 & 18.4~/~0.9 & 26.4~/~1.0 \\
        \bf GPT-Refine & \textbf{32.6}~/~1.0 & \textbf{54.5}~/~\textbf{0.8} & \textbf{35.6}~/~\textbf{0.9} & \textbf{24.4}~/~\textbf{1.0} & 22.5~/~1.3 & \textbf{7.5}~/~\textbf{0.9} & \textbf{24.0}~/~\textbf{0.9} & \textbf{28.7}~/~\textbf{1.0} \\
        \bf Self-Refine & 23.7~/~1.1 & 47.6~/~0.9 & 30.8~/~0.9 & 23.1~/~1.0 & 21.4~/~1.2 & 5.0~/~1.1 & 20.0~/~0.9 & 24.5~/~1.0 \\
        \hdashline
        \bf Context Refiner & 30.7~/~1.0 & 50.8~/~0.9 & 32.0~/~0.9 & 24.1~/~1.0 & 22.2~/~1.2 & 7.2~/~1.0 & 19.4~/~0.9 & 26.6~/~1.0 \\
        \bottomrule
    \end{tabular}}
    \caption{Performance (EM/ART) of context refinement methods for mitigating interference, measured by Exact Match (EM) and Average Retrieval Times (ART) across QA test sets. 
    }
    \label{table:refiner}
\end{table*}

\paragraph{Background}
Our preliminary findings and analysis in Sec.~\ref{ssec:components} demonstrate that we can marginally mitigate context interference by removing irrelevant information like previous documents and search queries in multi-turn search agents but not enough, which highlights the necessity of further capturing critical information and filtering noise in the latest retrieved documents (as \textbf{RQ ii}).
More generally, we desire to develop a context refiner $\boldsymbol{\mathcal{F}}$ that, in each round $i$, refine the context to preserve the most relevant information $\boldsymbol{\tilde d}_i=\boldsymbol{\mathcal{F}}(\boldsymbol{q}_{i-1},\boldsymbol{d}_i)$ to search query $\boldsymbol{q}_{i-1}$ from the latest retrieved documents $\boldsymbol{d}_i$.
$\boldsymbol{\tilde d}_i$ is served as the $i$-th observation $\boldsymbol{o}_i$ and then appended to $\boldsymbol{s}_i$ to generate $\boldsymbol{a}_i$.

Prior works on mitigating context interference are demonstrated in Appendix~\ref{ssec:context_interference}.
However, directly designing precise schemes to capture key information from numerous documents is challenging \citep{glass-etal-2022-re2g}. 
A compression model may focus on summarizing global contents, potentially leading to information loss or the introduction of extraneous knowledge \citep{li2025singleturnsurveymultiturninteractions}.
Relatively small LLMs exhibit limited capability for key information extraction (in Table~\ref{table:refiner}).
Therefore, we propose to distill a dataset for context refinement from advanced LLMs, enabling relatively weak models to refine context to mitigate interference.

\begin{figure}[!t]
    \centering
    \includegraphics[width=0.87\linewidth]{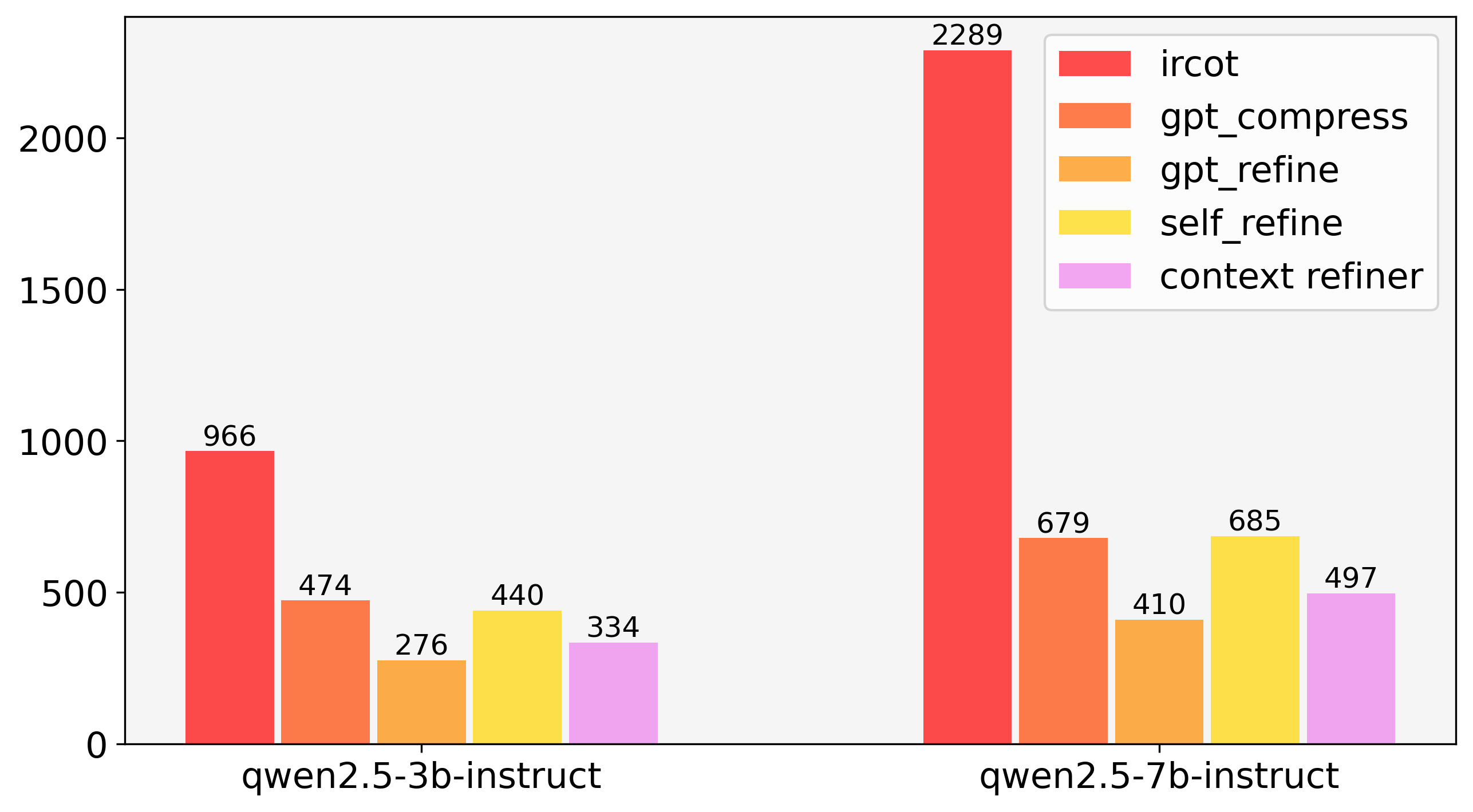}
    \caption{Averaged context lengths of search agents across all QA test sets, comparing IRCoT (baseline), three context refinement methods, and the proposed context refiner for efficiency assessments. 
    }
    \label{fig:context_len}
\end{figure}

\paragraph{Context Refiner}
Given a distill dataset $\boldsymbol{\mathcal{D}}_d=\{\boldsymbol{x}_i, \boldsymbol{\hat y}_i\}_{i=1}^N$, a foundation LLM $\boldsymbol{\mathcal{M}}_{\pi}$, an advanced teacher LLM $\boldsymbol{\mathcal{M}}_{T}$, a retriever $\boldsymbol{\mathcal{E}}$, and a knowledge base $\boldsymbol{K}_b$, we infer each query $\boldsymbol{x}$ on $\boldsymbol{\mathcal{M}}_{T}$ using IRCoT \citep{trivedi2023interleaving}.
In $i$-th round of inference for $\boldsymbol{x}$, the retriever return documents $\boldsymbol{d}_i =\boldsymbol{\mathcal{E}}(\boldsymbol{q}_{i-1}|\boldsymbol{K}_b)$.
Then we instruct the teacher model $\boldsymbol{\mathcal{M}}_{T}$ to specifically extract only critical information related to $\boldsymbol{q}_{i-1}$ from $\boldsymbol{d}_i$.
Extracted information $\boldsymbol{\tilde d}_i=\boldsymbol{\mathcal{M}}_{T}(\boldsymbol{q}_{i-1},\boldsymbol{d}_i)$ is then appended into context $\boldsymbol{s}_i$ to generate the next-step action $\boldsymbol{a}_i$.
For each correct trajectory with $\boldsymbol{y}\equiv\boldsymbol{\tilde{y}}$, we retain all step-wise pairs of extracted data $(\boldsymbol{\tilde d}_i, \text{<}\boldsymbol{d}_i, \boldsymbol{q}_{i-1}\text{>})$ and employ an entailment model to verify that $\boldsymbol{\tilde d}_i$ is entirely encompassed within $\boldsymbol{d}_i$ and does not introduce extra knowledge.
Finally, all qualified data points are re-formatted and incorporated into the context refinement dataset $\boldsymbol{\mathcal{D}}_{\text{c}}=\{ \boldsymbol{\tilde d}_j, \boldsymbol{d}_i, \boldsymbol{q}_{i}\}_{i=1}^{M}$.

Given $\boldsymbol{\mathcal{D}}_{\text{c}}$, we train the model $\boldsymbol{\mathcal{M}}_{\pi}$ to enable its ability of context refinement using supervised fine-tuning (SFT) as follows.
\begin{align}
    \pi^{*}&=\arg\min_{\pi}\mathcal{L}^{\text{SFT}}_{\pi}\\
    \mathcal{L}^{\text{SFT}}_{\pi} &= -\frac{1}{M}\sum_{i=1}^{M} 
\mathbb{E}_{(\boldsymbol{\tilde d}_i, \boldsymbol{d}_i, \boldsymbol{q}_{i})\sim \boldsymbol{\mathcal{D}}_{c}} \mathcal{L}^{(i)}_{\pi} \\
\mathcal{L}^{(i)}_{\pi} &= \log \boldsymbol{\mathcal{M}}_{\pi}(\boldsymbol{\tilde d}_i\mid \boldsymbol{d}_i,\boldsymbol{q}_{i})
\end{align}
$\boldsymbol{\mathcal{M}}_{\pi}$ is trained to generate refined documents $\boldsymbol{\tilde d}_i$, yielding a context refiner $\boldsymbol{\mathcal{F}}=\boldsymbol{\mathcal{M}}_{\pi^{*}}$ for dynamic context refinement in multi-turn search agents \footnote{The teacher model $\boldsymbol{\mathcal{M}}_{T}$ in this work is GPT-4 and the base model $\boldsymbol{\mathcal{M}}_{\pi}$ of context refiner $\boldsymbol{\mathcal{F}}$ is Qwen2.5-7b-Instruct or Qwen2.5-3b-Instruct, which is the same as their respective inference models.}.

\paragraph{Context Refinement Methods}
We employ several existing context refinement methods as comparisons:
1) We utilize the compression method by introducing GPT-4 \citep{gpt4} to summarize the previous contexts (\textbf{GPT-Compress});
2) We employ GPT-4 to dynamically refine the latest Top-$K$ retrieved documents $\boldsymbol{d}_{i}$ in each round based on the search query $\boldsymbol{q}_i$ (\textbf{GPT-Refine}).
3) We also employ the foundation LLM itself $\boldsymbol{\mathcal{M}}_{\pi}$ to dynamically refine the latest documents $\boldsymbol{d}_{i}$ using $\boldsymbol{q}_i$ (\textbf{Self-Refine}).

\paragraph{Results and Analysis}
As in Table~\ref{table:refiner} and Figure~\ref{fig:context_len}, \textbf{IRCoT} baseline is also presented for intuitive comparisons of context interference mitigation.
\textbf{GPT-Refine} consistently outperforms \textbf{GPT-Compression} in terms of reliability, while being slightly inferior in search times and context length, indicating that extracting and preserving search query-relevant key information is more effective in mitigating contextual interference than general summarization.
These two baselines can effectively reduce the retrieval wastes of search agents on irrelevant information in context over \textbf{IRCoT}.
\textbf{Self-Refine}, by contrast, does not yield performance improvements and even fails to reduce context length, which can be attributed to the lack of information extraction capability of the foundation model $\boldsymbol{\mathcal{M}}_{\pi}$.
In comparison, \textbf{Context Refiner} trained on the context refinement dataset achieves close performance to \textbf{GPT-Refine}, suggesting that even relatively weak LLMs can also acquire the ability to refine context through fine-tuning, without relying on external models for search agents. 
In addition, Context Refiner achieves marginally competitive performance to prompt-driven GPT-Refine which represents the performance of the teacher model. 
This suggests that Context Refiner can refine context to mitigate context interference and outperforms other compression and self-refine baselines.


\section{Context Refinement for Search Agent Training}
\label{sec:refinement}

\begin{figure*}[!ht]
    \centering
    \includegraphics[width=0.94\linewidth]{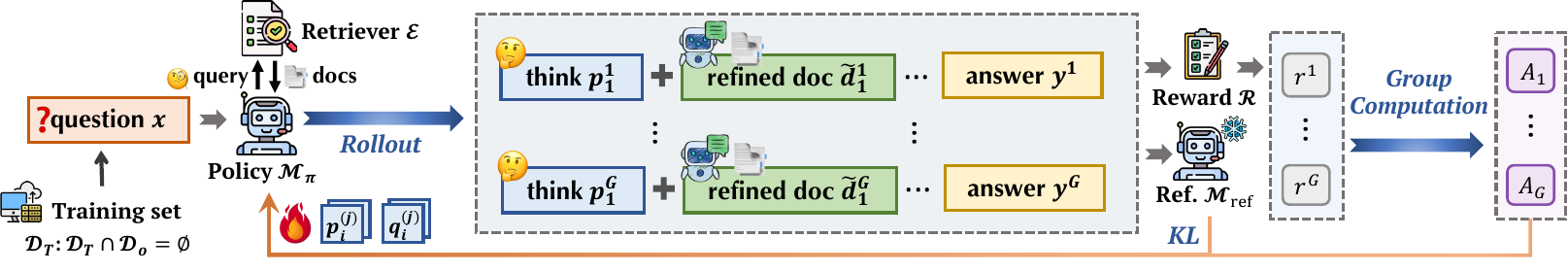}
    \caption{Demonstration of our proposed CRRL method.}
    \label{fig:train}
\end{figure*}

To further explore the potential of mitigating context interference in search agent training pipelines (as \textbf{RQ iii}), we extend our context refiner to the RL training pipeline of search agents, proposing a novel \textbf{C}ontext-\textbf{R}efined \textbf{R}einforcement \textbf{L}earning (CRRL) framework to dynamically refine context during training rollouts, reducing context interference in trajectory quality as follows.

\subsection{Context-Refined Reinforcement Learning}
\label{ssec:crpo}

Reinforcement learning (RL) \citep{kaelbling1996reinforcement} has emerged as a paradigm for search agent training \citep{song2025r1searcherincentivizingsearchcapability,jin2025search,chen2025learning} using PPO \citep{schulman2017proximal} and GRPO \citep{shao2024deepseekmath}.
During rollouts, these RL algorithms enable LLMs to iteratively interact with search engines and append the retrieved documents to contexts, potentially introducing interference in RL training and resulting in suboptimal performance of RL.
Nonetheless, current RL methods overlook the impact of context interference of rollouts.
Derived from Sec.~\ref{sec:interference}, we have figured out that removing previous documents and search queries, as well as leveraging context refiner for the latest documents in contexts, can eliminate context interference for multi-turn search agents, which can also improve the quality of rollouts.
Therefore, we propose the CRRL algorithm.

Current RL pipelines for search agents are mainly based on Proximal Policy Optimization (PPO)~\citep{schulman2017proximal} and Group Relative Policy Optimization (GRPO)~\citep{shao2024deepseekmath}.
GRPO performs multiple rollouts per task and calculates the relative reward within the group as the advantage, which is more lightweight without the value model and demonstrates comparable performance with PPO~\citep{jin2025search}.
Hence, this paper adopts GRPO as the default RL algorithm, and the proposed CRRL is also based on GRPO.

To mitigate context interference of GRPO rollouts and obtain high-quality trajectories, our CRRL dynamically refines the input context during rollout of multi-turn search agents with the context refiner $\boldsymbol{\mathcal{F}}$.
As shown in Figure~\ref{fig:train}, during rollout, the trajectories of CRRL only contain thinking steps and refined documents, which indicates that the action $\boldsymbol{a}^{j}_i=[\boldsymbol{p}^{j}_i, \boldsymbol{q}^{j}_i]$ is produced with the context $\boldsymbol{s}_i^{j}=[\boldsymbol{x}, \boldsymbol{p}^{j}_{0:i-1}, \boldsymbol{q}^{j}_{i-1}, \boldsymbol{\tilde d}^{j}_i]$ where $\boldsymbol{\tilde d}^{j}_i=\boldsymbol{\mathcal{F}}(\boldsymbol{d}_{i}^{j}), \boldsymbol{d}_{i}^{j}=\boldsymbol{\mathcal{E}}(\boldsymbol{q}_{i-1}^{j})$.
RL pipelines for search agents explicitly incorporate retrieval interleaved reasoning, and the token-level losses are only computed over the rollouts of LLM-generated tokens, including both search queries and thinking steps, where loss masking is introduced for retrieved tokens, ensuring the stabilization of training while preserving the ability to adaptively retrieve.
When optimizing the policy $\boldsymbol{\mathcal{M}}_{\pi}$ on $\boldsymbol{a}_i=[\boldsymbol{p}_i, \boldsymbol{q}_i]$, the CRRL algorithm can be represented as follows:

{

\begin{align}
     &\mathcal{L}^{\textrm{CRRL}}_{\boldsymbol{\mathcal{M}}_{\pi}}  = - \mathbb{E}_{\{ \boldsymbol a_i^{j} \}_{j=1}^G \sim {\boldsymbol{\mathcal{M}}}_{\text{ref}}( \cdot| \boldsymbol s_i^{j})} 
     \left [\mathcal{G}_{\pi} - \beta\text{KL} 
     \right ] \\
     &\mathcal{G}_{\pi} =  \frac{1}{G}\sum_{j=1}^G\frac{1}{\sum_{k=1}^{N-1}|{a}_{i,k}^j|}
     \left[ \mathcal{R}^j_{\pi} \right]\\
      &\mathcal{R}^j_{\pi,i,k}=\nonumber\\
     &\sum_{k}^{N-1} 
     \min \left( r_{\pi,i,k}^j A_j, \text{clip} \left( r_{\pi,i,k}^j, 1 - \epsilon, 1 + \epsilon \right) A_j \right)\\
      &r_{\pi,i,k}^j = \frac{{\boldsymbol{\mathcal{M}}}_\pi(a^j_{i,k}|\boldsymbol x, \boldsymbol a^j_{i,<k},\boldsymbol{s}_i^j)}{\boldsymbol{\mathcal{M}}_{{\mathrm{ref}}}(a^j_{i,k}|\boldsymbol x, \boldsymbol a^j_{i,<k},\boldsymbol{s}_i^j)}
\end{align}
}
where $\boldsymbol{\mathcal{M}}_{{\mathrm{ref}}}$ represents reference model. 
The term $\epsilon$ is a clipping ratio.
$\beta$ is the coefficient for the KL divergence.
The advantage estimate is computed on the group-relative rewards of trajectory $\boldsymbol{\tau}^j$ as
\begin{align}
    A_j = \frac{r^j-\mu^j}{\sigma^j}
\end{align}
where $r^j=\boldsymbol{\mathcal{R}}(\boldsymbol{y}^j)$ is the final reward, and $\mu^j$ and $\sigma^j$ denote the mean and standard deviation of the rewards within the group.

\begin{table*}[!t]
    \centering
    \footnotesize
    {
    \begin{tabular}{lcccccccccccc}
        \toprule
        \multirow{2}{*}{\textbf{Methods}} & \multicolumn{3}{c}{\textbf{Single-Hop QA}} & \multicolumn{4}{c}{\textbf{Multi-Hop QA}} & \multirow{2}{*}{\textit{\textbf{Avg.}}} \\
        \cmidrule(r){2-4}\cmidrule(r){5-8}
         & \textbf{NQ} & \textbf{TriviaQA} & \textbf{PopQA} & \textbf{HotpotQA} & \textbf{2wiki} & \textbf{Musique} & \textbf{Bamboogle} & \\
        \hline
        \hline
        \rowcolor{seashell}
        \multicolumn{9}{c}{\textbf{{Qwen2.5-7b-Instruct}}} \\
        \hline
        \bf IRCoT & 30.6~/~2.0 & 51.2~/~1.8 & 31.7~/~2.1 & 24.6~/~3.0 & 18.0~/~3.4 & 9.8~/~3.1 & 28.8~/~2.5 & 27.5~/~2.6 \\
        \hdashline
        \bf SFT & 26.7~/~0.0 & 40.5~/~0.0 & 15.3~/~0.0 & 22.4~/~0.0 & 20.5~/~0.0 & 6.3~/~0.0 & 11.2~/~0.0 & 20.4~/~0.0 \\
        \bf R1 & 24.7~/~0.0 & 55.4~/~0.0 & 20.0~/~0.0 & 19.4~/~0.0 & 17.5~/~0.0 & 7.8~/~0.0 & 22.4~/~0.0 & 23.9~/~0.0 \\
        \bf RFT & 34.4~/~2.2 & 57.5~/~2.0 & 36.3~/~2.4 & 31.8~/~3.0 & 26.5~/~3.6 & 11.5~/~3.2 & 33.6~/~2.7 & 33.1~/~2.7 \\
        \bf Search-GRPO & 35.5~/~1.7 & 58.4~/~1.4 & 38.5~/~1.8 & 35.6~/~2.7 & 27.0~/~3.0 & 11.5~/~2.3 & 36.0~/~1.8 & 34.6~/~2.1 \\
        \bf Search-o1 & 37.5~/~1.6 & 60.4~/~1.4 & 39.2~/~1.8 & 37.8~/~2.3 & 29.6~/~2.6 & 13.4~/~2.0 & 36.8~/~1.6 & 36.2~/~1.9 \\
        \hdashline
        \bf CRRL & \bf 38.1~/~\bf 1.4 & \bf 60.1~/~\bf 1.2 & \bf 39.4~/~\bf 1.6 & \bf 38.2~/~\bf 2.0 & \bf 30.5~/~\bf 2.5 & \bf 13.2~/~\bf 1.9 & \bf 36.8~/~\bf 1.5 & \bf 36.6~/~\bf 1.7 \\
        \hline
        \rowcolor{seashell}
        \multicolumn{9}{c}{\textbf{{Qwen2.5-3b-Instruct}}} \\
        \hline
        \bf IRCoT & 21.6~/~1.2 & 45.2~/~1.1 & 29.6~/~1.1 & 24.1~/~1.6 & 23.5~/~1.9 & 6.8~/~1.6 & 19.2~/~1.4 & 24.3~/~1.4 \\
        \hdashline
        \bf SFT & 23.8~/~0.0 & 36.2~/~0.0 & 11.4~/~0.0 & 18.6~/~0.0 & 22.8~/~0.0 & 4.8~/~0.0 & 8.0~/~0.0 & 17.9~/~0.0 \\
        \bf R1 & 19.6~/~0.0 & 42.9~/~0.0 & 15.2~/~0.0 & 18.5~/~0.0 & 26.5~/~0.0 & 8.0~/~0.0 & 19.2~/~0.0 & 21.4~/~0.0 \\
        \bf RFT & 29.4~/~1.3 & 47.5~/~1.2 & 32.5~/~1.1 & 26.0~/~1.8 & 24.4~/~2.2 & 6.4~/~1.8 & 21.0~/~1.8 & 26.7~/~1.6 \\
        \bf Search-GRPO & 33.7~/~1.4 & 52.1~/~1.1 & 34.2~/~1.1 & 27.1~/~1.6 & 26.5~/~1.6 & 11.8~/~1.7 & {\bf 23.2}~/~1.5 & 29.8~/~1.4 \\
        \bf Search-o1 & 34.7~/~1.1 & 53.2~/~1.1 & 33.5~/~1.1 & 27.0~/~\bf 1.3 & 28.6~/~1.5 & 12.0~/~1.6 & 23.2~/~\bf 1.3 & 30.3~/~\bf 1.3 \\
        \hdashline
        \bf CRRL & \bf 34.7~/~\bf 1.1 & \bf 54.5~/~\bf 1.1 & 3\bf 6.8~/~\bf 1.0 & \bf 29.0~/~1.4 & \bf 30.0~/~\bf 1.5 & \bf 12.4~/~\bf 1.6 & \bf 23.2~/~1.4 & \bf 31.5~/~\bf 1.3 \\
        \bottomrule
    \end{tabular}}
    \caption{Performance results of EM/ART across QA test sets on several baselines as well as our proposed CRRL method, measured by Exact Match (EM) and Average Retrieval Times (ART).}
    \label{table:training}
\end{table*}

\subsection{Training Setup}
\label{ssec:exp_setup}

We sample 40k data from the training sets of NQ and HotpotQA and merge them into $\boldsymbol{\mathcal{D}}_t$ where $\boldsymbol{\mathcal{D}}_t\cap \ \boldsymbol{\mathcal{D}}_d= \varnothing$.
Evaluation is conducted on seven QA datasets to assess both in-domain and out-of-domain performance.
The GRPO implementation is based on Verl \citep{sheng2025verl}.
More implementation details can be found in Appendix~\ref{appendix:implementation}.


We employ two types of baselines as comparisons.
1) \textbf{Retrieval-free Fine-Tuning}:
We train LLMs using both supervised fine-tuning \textbf{(SFT)} and \textbf{(GRPO)}-based RL \citep{shao2024deepseekmath} methods which only contain reasoning and answer steps.
2) \textbf{Retrieval-based Fine-Tuning}:
To obtain the trajectories with the retriever, we utilize rejection sampling to generate several candidate responses by interacting with the search engine for each question from the training set $\boldsymbol{\mathcal{D}}_t$.
Then we collect paths including correct answers for rejection fine-tuning \textbf{(RFT)} \citep{yuan2023scalingrelationshiplearningmathematical}.
We also employ the RL pipeline for multi-turn search agents in \textbf{Search-GRPO} following \citep{jin2025search} and \textbf{Search-o1} following \citep{li-etal-2025-search}, which optimizes LLM rollouts to autonomously call the retriever when they lack relevant knowledge.

\subsection{Results and Analysis}
\label{sec:analysis}

\begin{table}[!ht]
    \centering
    \footnotesize
    {
    \begin{tabular}{lcccccccccccc}
        \toprule
        {\textbf{LLMs}} & {\textbf{Methods}} & {\textbf{ART}} & {\textbf{Len.}} & {{\textbf{AIT}}} \\
        \hline
        \hline
        \multirow{3}{*}{\textbf{{Qwen2.5-7b}}} & \bf IRCoT & 2.6 & 2.3k & 22.4s \\
        & \bf Search-GRPO & 2.1 & 0.9k & 17.7s \\
        & \bf CRRL & 1.6 & 0.7k & 16.9s \\
        \hline
        \multirow{3}{*}{\textbf{{Qwen2.5-3b}}} & \bf IRCoT & 1.4 & 0.9k & 7.5s \\
        & \bf Search-GRPO & 1.4 & 0.7k & 4.5s \\
        & \bf CRRL & 1.2 & 0.5k & 4.2s \\
        \bottomrule
    \end{tabular}}
    \caption{Efficiency comparisons of baselines and our proposed CRRL during inference on test sets of average retrieval times (ART), average context length (Len.), and average inference time per question (AIT).}
    \label{table:efficiency}
\end{table}

In Table~\ref{table:training}, experimental results demonstrate that our proposed CRRL method outperforms other baselines in EM scores, suggesting that incorporating context refinement into the RL training pipeline of search agents can effectively mitigate context interference and improve the reliability of generations.
In Table~\ref{table:efficiency}, during the inference phase, our proposed CRRL achieves a significant reduction in both ART and Len. compared to IRCoT and Search-GRPO, which compensate for context refiner overhead arising from the extra inference costs, and result in lower AIT, suggesting that introducing a context refiner during RL training to relax contextual interference can finally improve the efficiency of the search agent in inference.

\section{Conclusion}

This work investigates the context interference issue on search agents across a variety of QA benchmarks.
We first demonstrate that context interference, largely stemming from the latest retrieved documents, poses a key challenge to multi-turn search agents.
Then, we present a distill-based context refiner to dynamically mitigate interference in multi-turn search agents and thus significantly boost both reliability and efficiency. 
Furthermore, we introduce context refinement into the RL training pipelines of search agents, which can further yield performance improvements.
These findings highlight the importance of context refinement to mitigate context interference to construct reliable and efficient search agents, paving the way for a new paradigm of dynamic ``refine context and then generate'' for AI agents in future work.

\section*{Limitations}

The limitations and future work of this study are listed as follows:

\paragraph{Task Settings}
This study mainly focuses on mitigating the context interference issue on search agent tasks.
However, similar problems also arise in other agent settings, such as \textbf{tool use} and \textbf{planning}.
The principal factors of context interference in various tasks may differ, necessitating specific mitigation strategies to further improve reliability and efficiency of AI agents.

\paragraph{Paradigm Design}
The context refiner proposed in this study is implemented as an auxiliary module to the search agent, rather than being integrated into the training pipelines of agents.
In future work, we plan to develop a dedicated training algorithm that internalizes the context refinement capability within agents themselves.
This may inspire a new paradigm of ``receiving observation $\rightarrow$ refining context $\rightarrow$ generating action'' for agents, which can dynamically eliminate the context interference, achieving more efficient and reliable AI agents.


\section*{Acknowledgments}

This work is partially supported by Hong Kong RGC GRF No. 14206324.


\bibliography{custom}

@misc{jia2025fastslowtoolaugmentedthinking,
      title={Fast, Slow, and Tool-augmented Thinking for LLMs: A Review}, 
      author={Xinda Jia and Jinpeng Li and Zezhong Wang and Jingjing Li and Xingshan Zeng and Yasheng Wang and Weinan Zhang and Yong Yu and Weiwen Liu},
      year={2025},
      eprint={2508.12265},
      archivePrefix={arXiv},
      primaryClass={cs.CL},
      url={https://arxiv.org/abs/2508.12265}, 
}

@misc{laban2025llmslostmultiturnconversation,
      title={LLMs Get Lost In Multi-Turn Conversation}, 
      author={Philippe Laban and Hiroaki Hayashi and Yingbo Zhou and Jennifer Neville},
      year={2025},
      eprint={2505.06120},
      archivePrefix={arXiv},
      primaryClass={cs.CL},
      url={https://arxiv.org/abs/2505.06120}, 
}

@misc{li2025singleturnsurveymultiturninteractions,
      title={Beyond Single-Turn: A Survey on Multi-Turn Interactions with Large Language Models}, 
      author={Yubo Li and Xiaobin Shen and Xinyu Yao and Xueying Ding and Yidi Miao and Ramayya Krishnan and Rema Padman},
      year={2025},
      eprint={2504.04717},
      archivePrefix={arXiv},
      primaryClass={cs.CL},
      url={https://arxiv.org/abs/2504.04717}, 
}

@misc{jiang2025enhancingrobustnesslargelanguage,
      title={Enhancing Robustness in Large Language Models: Prompting for Mitigating the Impact of Irrelevant Information}, 
      author={Ming Jiang and Tingting Huang and Biao Guo and Yao Lu and Feng Zhang},
      year={2025},
      eprint={2408.10615},
      archivePrefix={arXiv},
      primaryClass={cs.CL},
      url={https://arxiv.org/abs/2408.10615}, 
}

@misc{nguyen2025maragmultiagentretrievalaugmentedgeneration,
      title={MA-RAG: Multi-Agent Retrieval-Augmented Generation via Collaborative Chain-of-Thought Reasoning}, 
      author={Thang Nguyen and Peter Chin and Yu-Wing Tai},
      year={2025},
      eprint={2505.20096},
      archivePrefix={arXiv},
      primaryClass={cs.CL},
      url={https://arxiv.org/abs/2505.20096}, 
}

@misc{yu2024rankragunifyingcontextranking,
      title={RankRAG: Unifying Context Ranking with Retrieval-Augmented Generation in LLMs}, 
      author={Yue Yu and Wei Ping and Zihan Liu and Boxin Wang and Jiaxuan You and Chao Zhang and Mohammad Shoeybi and Bryan Catanzaro},
      year={2024},
      eprint={2407.02485},
      archivePrefix={arXiv},
      primaryClass={cs.CL},
      url={https://arxiv.org/abs/2407.02485}, 
}

@misc{haseeb2025contextengineeringmultiagentllm,
      title={Context Engineering for Multi-Agent LLM Code Assistants Using Elicit, NotebookLM, ChatGPT, and Claude Code}, 
      author={Muhammad Haseeb},
      year={2025},
      eprint={2508.08322},
      archivePrefix={arXiv},
      primaryClass={cs.SE},
      url={https://arxiv.org/abs/2508.08322}, 
}

@misc{yi2025surveyrecentadvancesllmbased,
      title={A Survey on Recent Advances in LLM-Based Multi-turn Dialogue Systems}, 
      author={Zihao Yi and Jiarui Ouyang and Zhe Xu and Yuwen Liu and Tianhao Liao and Haohao Luo and Ying Shen},
      year={2025},
      eprint={2402.18013},
      archivePrefix={arXiv},
      primaryClass={cs.CL},
      url={https://arxiv.org/abs/2402.18013}, 
}

@misc{shao2024deepseekmath,
  author = {Zhihong Shao and Peiyi Wang and Qihao Zhu, Runxin Xu and Junxiao Song and Mingchuan Zhang and Y.K. Li and Y. Wu and Daya Guo},
  title = {DeepSeekMath: Pushing the Limits of Mathematical Reasoning in Open Language Models},
  journal = {CoRR},
  volume = {abs/2402.03300},
  year = {2024},
  url = {https://arxiv.org/abs/2402.03300},
}

@misc{deepseekai2025deepseekr1incentivizingreasoningcapability,
      title={DeepSeek-R1: Incentivizing Reasoning Capability in LLMs via Reinforcement Learning}, 
      author={DeepSeek-AI},
      year={2025},
      eprint={2501.12948},
      archivePrefix={arXiv},
      primaryClass={cs.CL},
      url={https://arxiv.org/abs/2501.12948}, 
}

@misc{gpt5,
  author = {OpenAI},
  title = {Introducing GPT-5
},
  year = 2025,
  url = {https://openai.com/index/introducing-gpt-5/},
  note = {https://openai.com/index/introducing-gpt-5/},
  urldate = {2024-05}
}

@misc{gpt4,
  author = {OpenAI},
  title = {GPT-4
},
  year = 2023,
  url = {https://openai.com/index/gpt-4-research/},
  note = {https://openai.com/index/gpt-4-research/},
  urldate = {2023-03}
}

@misc{jin2025search,
      title={Search-R1: Training LLMs to Reason and Leverage Search Engines with Reinforcement Learning}, 
      author={Bowen Jin and Hansi Zeng and Zhenrui Yue and Jinsung Yoon and Sercan Arik and Dong Wang and Hamed Zamani and Jiawei Han},
      year={2025},
      eprint={2503.09516},
      archivePrefix={arXiv},
      primaryClass={cs.CL},
      url={https://arxiv.org/abs/2503.09516}, 
}

@misc{song2025r1searcherincentivizingsearchcapability,
      title={R1-Searcher: Incentivizing the Search Capability in LLMs via Reinforcement Learning}, 
      author={Huatong Song and Jinhao Jiang and Yingqian Min and Jie Chen and Zhipeng Chen and Wayne Xin Zhao and Lei Fang and Ji-Rong Wen},
      year={2025},
      eprint={2503.05592},
      archivePrefix={arXiv},
      primaryClass={cs.AI},
      url={https://arxiv.org/abs/2503.05592}, 
}

@inproceedings{dong2025understand,
author = {Dong, Guanting and Zhu, Yutao and Zhang, Chenghao and Wang, Zechen and Wen, Ji-Rong and Dou, Zhicheng},
title = {Understand What LLM Needs: Dual Preference Alignment for Retrieval-Augmented Generation},
year = {2025},
isbn = {9798400712746},
publisher = {Association for Computing Machinery},
address = {New York, NY, USA},
url = {https://doi.org/10.1145/3696410.3714717},
doi = {10.1145/3696410.3714717},
booktitle = {Proceedings of the ACM on Web Conference 2025},
pages = {4206–4225},
numpages = {20},
location = {Sydney NSW, Australia},
series = {WWW '25}
}

@misc{coleman2023incontextinterferencechatbasedlarge,
      title={In-context Interference in Chat-based Large Language Models}, 
      author={Eric Nuertey Coleman and Julio Hurtado and Vincenzo Lomonaco},
      year={2023},
      eprint={2309.12727},
      archivePrefix={arXiv},
      primaryClass={cs.AI},
      url={https://arxiv.org/abs/2309.12727}, 
}

@inproceedings{gupta-etal-2024-llm,
    title = "{LLM} Task Interference: An Initial Study on the Impact of Task-Switch in Conversational History",
    author = "Gupta, Akash  and
      Sheth, Ivaxi  and
      Raina, Vyas  and
      Gales, Mark  and
      Fritz, Mario",
    editor = "Al-Onaizan, Yaser  and
      Bansal, Mohit  and
      Chen, Yun-Nung",
    booktitle = "Proceedings of the 2024 Conference on Empirical Methods in Natural Language Processing",
    month = nov,
    year = "2024",
    address = "Miami, Florida, USA",
    publisher = "Association for Computational Linguistics",
    url = "https://aclanthology.org/2024.emnlp-main.811/",
    doi = "10.18653/v1/2024.emnlp-main.811",
    pages = "14633--14652"
}

@article{peng2023study,
  title={A study of generative large language model for medical research and healthcare},
  author={Peng, Cheng and Yang, Xi and Chen, Aokun and Smith, Kaleb E and PourNejatian, Nima and Costa, Anthony B and Martin, Cheryl and Flores, Mona G and Zhang, Ying and Magoc, Tanja and others},
  journal={NPJ digital medicine},
  volume={6},
  number={1},
  pages={210},
  year={2023},
  publisher={Nature Publishing Group UK London}
}

@inproceedings{li2023large,
author = {Li, Yinheng and Wang, Shaofei and Ding, Han and Chen, Hang},
title = {Large Language Models in Finance: A Survey},
year = {2023},
isbn = {9798400702402},
publisher = {Association for Computing Machinery},
address = {New York, NY, USA},
url = {https://doi.org/10.1145/3604237.3626869},
doi = {10.1145/3604237.3626869},
booktitle = {Proceedings of the Fourth ACM International Conference on AI in Finance},
pages = {374–382},
numpages = {9},
location = {Brooklyn, NY, USA},
series = {ICAIF '23}
}

@article{zhao2024dense,
  title={Dense text retrieval based on pretrained language models: A survey},
  author={Zhao, Wayne Xin and Liu, Jing and Ren, Ruiyang and Wen, Ji-Rong},
  journal={ACM Transactions on Information Systems},
  volume={42},
  number={4},
  pages={1--60},
  year={2024},
  publisher={ACM New York, NY}
}

@misc{gao2024retrievalaugmentedgenerationlargelanguage,
      title={Retrieval-Augmented Generation for Large Language Models: A Survey}, 
      author={Yunfan Gao and Yun Xiong and Xinyu Gao and Kangxiang Jia and Jinliu Pan and Yuxi Bi and Yi Dai and Jiawei Sun and Meng Wang and Haofen Wang},
      year={2024},
      eprint={2312.10997},
      archivePrefix={arXiv},
      primaryClass={cs.CL},
      url={https://arxiv.org/abs/2312.10997}, 
}

@article{lewis2020retrieval,
  title={Retrieval-augmented generation for knowledge-intensive nlp tasks},
  author={Lewis, Patrick and Perez, Ethan and Piktus, Aleksandra and Petroni, Fabio and Karpukhin, Vladimir and Goyal, Naman and K{\"u}ttler, Heinrich and Lewis, Mike and Yih, Wen-tau and Rockt{\"a}schel, Tim and others},
  journal={Advances in neural information processing systems},
  volume={33},
  pages={9459--9474},
  year={2020}
}

@misc{xiong2025rag,
      title={RAG-Gym: Systematic Optimization of Language Agents for Retrieval-Augmented Generation}, 
      author={Guangzhi Xiong and Qiao Jin and Xiao Wang and Yin Fang and Haolin Liu and Yifan Yang and Fangyuan Chen and Zhixing Song and Dengyu Wang and Minjia Zhang and Zhiyong Lu and Aidong Zhang},
      year={2025},
      eprint={2502.13957},
      archivePrefix={arXiv},
      primaryClass={cs.CL},
      url={https://arxiv.org/abs/2502.13957}, 
}

@article{schick2023toolformer,
  title={Toolformer: Language models can teach themselves to use tools},
  author={Schick, Timo and Dwivedi-Yu, Jane and Dess{\`\i}, Roberto and Raileanu, Roberta and Lomeli, Maria and Hambro, Eric and Zettlemoyer, Luke and Cancedda, Nicola and Scialom, Thomas},
  journal={Advances in Neural Information Processing Systems},
  volume={36},
  pages={68539--68551},
  year={2023}
}

@inproceedings{trivedi2023interleaving,
    title = "Interleaving Retrieval with Chain-of-Thought Reasoning for Knowledge-Intensive Multi-Step Questions",
    author = "Trivedi, Harsh  and
      Balasubramanian, Niranjan  and
      Khot, Tushar  and
      Sabharwal, Ashish",
    editor = "Rogers, Anna  and
      Boyd-Graber, Jordan  and
      Okazaki, Naoaki",
    booktitle = "Proceedings of the 61st Annual Meeting of the Association for Computational Linguistics (Volume 1: Long Papers)",
    month = jul,
    year = "2023",
    address = "Toronto, Canada",
    publisher = "Association for Computational Linguistics",
    url = "https://aclanthology.org/2023.acl-long.557/",
    doi = "10.18653/v1/2023.acl-long.557",
    pages = "10014--10037"
}

@inproceedings{yaoreact,
place = {Country unknown/Code not available}, 
title = {ReAct: Synergizing Reasoning and Acting in Language Models}, 
url = {https://par.nsf.gov/biblio/10451467}, 
abstractNote = {While large language models (LLMs) have demonstrated impressive capabilities across tasks in language understanding and interactive decision making, their abilities for reasoning (e.g. chain-of-thought prompting) and acting (e.g. action plan generation) have primarily been studied as separate topics. In this paper, we explore the use of LLMs to generate both reasoning traces and task-specific actions in an interleaved manner, allowing for greater synergy between the two: reasoning traces help the model induce, track, and update action plans as well as handle exceptions, while actions allow it to interface with external sources, such as knowledge bases or environments, to gather additional information. We apply our approach, named ReAct, to a diverse set of language and decision making tasks and demonstrate its effectiveness over state-of-the-art baselines, as well as improved human interpretability and trustworthiness over methods without reasoning or acting components. Concretely, on question answering (HotpotQA) and fact verification (Fever), ReAct overcomes issues of hallucination and error propagation prevalent in chain-of-thought reasoning by interacting with a simple Wikipedia API, and generates human-like task-solving trajectories that are more interpretable than baselines without reasoning traces. On two interactive decision making benchmarks (ALFWorld and WebShop), ReAct outperforms imitation and reinforcement learning methods by an absolute success rate of 34% and 10% respectively, while being prompted with only one or two in-context examples.},
booktitle = {ICLR}, 
author = {Yao, Shunyu and Zhao, Jeffrey and Yu, Dian and Du, Nan and Shafran, Izhak and Narasimhan, Karthik and Cao, Yuan},
publisher= {The Eleventh International Conference on Learning Representations},
address={Kigali Rwanda},
year={2023},
pages={no}
}

@article{kaelbling1996reinforcement,
  title={Reinforcement learning: A survey},
  author={Kaelbling, Leslie Pack and Littman, Michael L and Moore, Andrew W},
  journal={Journal of artificial intelligence research},
  volume={4},
  pages={237--285},
  year={1996}
}

@article{ouyang2022training,
  title={Training language models to follow instructions with human feedback},
  author={Ouyang, Long and Wu, Jeffrey and Jiang, Xu and Almeida, Diogo and Wainwright, Carroll and Mishkin, Pamela and Zhang, Chong and Agarwal, Sandhini and Slama, Katarina and Ray, Alex and others},
  journal={Advances in neural information processing systems},
  volume={35},
  pages={27730--27744},
  year={2022}
}

@misc{chen2025learning,
      title={ReSearch: Learning to Reason with Search for LLMs via Reinforcement Learning}, 
      author={Mingyang Chen and Linzhuang Sun and Tianpeng Li and Haoze Sun and Yijie Zhou and Chenzheng Zhu and Haofen Wang and Jeff Z. Pan and Wen Zhang and Huajun Chen and Fan Yang and Zenan Zhou and Weipeng Chen},
      year={2025},
      eprint={2503.19470},
      archivePrefix={arXiv},
      primaryClass={cs.AI},
      url={https://arxiv.org/abs/2503.19470}, 
}

@inproceedings{jacqmin-etal-2022-follow,
    title = "``Do you follow me?'': A Survey of Recent Approaches in Dialogue State Tracking",
    author = "Jacqmin, L{\'e}o  and
      Rojas Barahona, Lina M.  and
      Favre, Benoit",
    editor = "Lemon, Oliver  and
      Hakkani-Tur, Dilek  and
      Li, Junyi Jessy  and
      Ashrafzadeh, Arash  and
      Garcia, Daniel Hern{\'a}ndez  and
      Alikhani, Malihe  and
      Vandyke, David  and
      Du{\v{s}}ek, Ond{\v{r}}ej",
    booktitle = "Proceedings of the 23rd Annual Meeting of the Special Interest Group on Discourse and Dialogue",
    month = sep,
    year = "2022",
    address = "Edinburgh, UK",
    publisher = "Association for Computational Linguistics",
    url = "https://aclanthology.org/2022.sigdial-1.33/",
    doi = "10.18653/v1/2022.sigdial-1.33",
    pages = "336--350"
}

@inproceedings{glass-etal-2022-re2g,
    title = "{R}e2{G}: Retrieve, Rerank, Generate",
    author = "Glass, Michael  and
      Rossiello, Gaetano  and
      Chowdhury, Md Faisal Mahbub  and
      Naik, Ankita  and
      Cai, Pengshan  and
      Gliozzo, Alfio",
    editor = "Carpuat, Marine  and
      de Marneffe, Marie-Catherine  and
      Meza Ruiz, Ivan Vladimir",
    booktitle = "Proceedings of the 2022 Conference of the North American Chapter of the Association for Computational Linguistics: Human Language Technologies",
    month = jul,
    year = "2022",
    address = "Seattle, United States",
    publisher = "Association for Computational Linguistics",
    url = "https://aclanthology.org/2022.naacl-main.194/",
    doi = "10.18653/v1/2022.naacl-main.194",
    pages = "2701--2715"
}

@inproceedings{jiang2023llmlingua,
    title = "{LLML}ingua: Compressing Prompts for Accelerated Inference of Large Language Models",
    author = "Jiang, Huiqiang  and
      Wu, Qianhui  and
      Lin, Chin-Yew  and
      Yang, Yuqing  and
      Qiu, Lili",
    editor = "Bouamor, Houda  and
      Pino, Juan  and
      Bali, Kalika",
    booktitle = "Proceedings of the 2023 Conference on Empirical Methods in Natural Language Processing",
    month = dec,
    year = "2023",
    address = "Singapore",
    publisher = "Association for Computational Linguistics",
    url = "https://aclanthology.org/2023.emnlp-main.825/",
    doi = "10.18653/v1/2023.emnlp-main.825",
    pages = "13358--13376"
}

@misc{rajeev2025catsconfusereasoningllm,
      title={Cats Confuse Reasoning LLM: Query Agnostic Adversarial Triggers for Reasoning Models}, 
      author={Meghana Rajeev and Rajkumar Ramamurthy and Prapti Trivedi and Vikas Yadav and Oluwanifemi Bamgbose and Sathwik Tejaswi Madhusudan and James Zou and Nazneen Rajani},
      year={2025},
      eprint={2503.01781},
      archivePrefix={arXiv},
      primaryClass={cs.CL},
      url={https://arxiv.org/abs/2503.01781}, 
}

@inproceedings{karpukhin2020dense,
    title = "Dense Passage Retrieval for Open-Domain Question Answering",
    author = "Karpukhin, Vladimir  and
      Oguz, Barlas  and
      Min, Sewon  and
      Lewis, Patrick  and
      Wu, Ledell  and
      Edunov, Sergey  and
      Chen, Danqi  and
      Yih, Wen-tau",
    editor = "Webber, Bonnie  and
      Cohn, Trevor  and
      He, Yulan  and
      Liu, Yang",
    booktitle = "Proceedings of the 2020 Conference on Empirical Methods in Natural Language Processing (EMNLP)",
    month = nov,
    year = "2020",
    address = "Online",
    publisher = "Association for Computational Linguistics",
    url = "https://aclanthology.org/2020.emnlp-main.550/",
    doi = "10.18653/v1/2020.emnlp-main.550",
    pages = "6769--6781"
}

@article{kwiatkowski2019natural,
  title={Natural questions: a benchmark for question answering research},
  author={Kwiatkowski, Tom and Palomaki, Jennimaria and Redfield, Olivia and Collins, Michael and Parikh, Ankur and Alberti, Chris and Epstein, Danielle and Polosukhin, Illia and Devlin, Jacob and Lee, Kenton and others},
  journal={Transactions of the Association for Computational Linguistics},
  volume={7},
  pages={453--466},
  year={2019},
  publisher={MIT Press One Rogers Street, Cambridge, MA 02142-1209, USA journals-info~…}
}

@inproceedings{joshi2017triviaqa,
    title = "{T}rivia{QA}: A Large Scale Distantly Supervised Challenge Dataset for Reading Comprehension",
    author = "Joshi, Mandar  and
      Choi, Eunsol  and
      Weld, Daniel  and
      Zettlemoyer, Luke",
    editor = "Barzilay, Regina  and
      Kan, Min-Yen",
    booktitle = "Proceedings of the 55th Annual Meeting of the Association for Computational Linguistics (Volume 1: Long Papers)",
    month = jul,
    year = "2017",
    address = "Vancouver, Canada",
    publisher = "Association for Computational Linguistics",
    url = "https://aclanthology.org/P17-1147/",
    doi = "10.18653/v1/P17-1147",
    pages = "1601--1611"
}

@inproceedings{mallen2022not,
    title = "When Not to Trust Language Models: Investigating Effectiveness of Parametric and Non-Parametric Memories",
    author = "Mallen, Alex  and
      Asai, Akari  and
      Zhong, Victor  and
      Das, Rajarshi  and
      Khashabi, Daniel  and
      Hajishirzi, Hannaneh",
    editor = "Rogers, Anna  and
      Boyd-Graber, Jordan  and
      Okazaki, Naoaki",
    booktitle = "Proceedings of the 61st Annual Meeting of the Association for Computational Linguistics (Volume 1: Long Papers)",
    month = jul,
    year = "2023",
    address = "Toronto, Canada",
    publisher = "Association for Computational Linguistics",
    url = "https://aclanthology.org/2023.acl-long.546/",
    doi = "10.18653/v1/2023.acl-long.546",
    pages = "9802--9822"
}

@inproceedings{yang2018hotpotqa,
    title = "{H}otpot{QA}: A Dataset for Diverse, Explainable Multi-hop Question Answering",
    author = "Yang, Zhilin  and
      Qi, Peng  and
      Zhang, Saizheng  and
      Bengio, Yoshua  and
      Cohen, William  and
      Salakhutdinov, Ruslan  and
      Manning, Christopher D.",
    editor = "Riloff, Ellen  and
      Chiang, David  and
      Hockenmaier, Julia  and
      Tsujii, Jun{'}ichi",
    booktitle = "Proceedings of the 2018 Conference on Empirical Methods in Natural Language Processing",
    month = oct # "-" # nov,
    year = "2018",
    address = "Brussels, Belgium",
    publisher = "Association for Computational Linguistics",
    url = "https://aclanthology.org/D18-1259/",
    doi = "10.18653/v1/D18-1259",
    pages = "2369--2380"
}

@inproceedings{ho2020constructing,
    title = "Constructing A Multi-hop {QA} Dataset for Comprehensive Evaluation of Reasoning Steps",
    author = "Ho, Xanh  and
      Duong Nguyen, Anh-Khoa  and
      Sugawara, Saku  and
      Aizawa, Akiko",
    editor = "Scott, Donia  and
      Bel, Nuria  and
      Zong, Chengqing",
    booktitle = "Proceedings of the 28th International Conference on Computational Linguistics",
    month = dec,
    year = "2020",
    address = "Barcelona, Spain (Online)",
    publisher = "International Committee on Computational Linguistics",
    url = "https://aclanthology.org/2020.coling-main.580/",
    doi = "10.18653/v1/2020.coling-main.580",
    pages = "6609--6625"
}

@article{trivedi2022musique,
  title={MuSiQue: Multihop Questions via Single-hop Question Composition},
  author={Trivedi, Harsh and Balasubramanian, Niranjan and Khot, Tushar and Sabharwal, Ashish},
  journal={Transactions of the Association for Computational Linguistics},
  volume={10},
  pages={539--554},
  year={2022},
  publisher={MIT Press One Broadway, 12th Floor, Cambridge, Massachusetts 02142, USA~…}
}

@inproceedings{press2022measuring,
    title = "Measuring and Narrowing the Compositionality Gap in Language Models",
    author = "Press, Ofir  and
      Zhang, Muru  and
      Min, Sewon  and
      Schmidt, Ludwig  and
      Smith, Noah  and
      Lewis, Mike",
    editor = "Bouamor, Houda  and
      Pino, Juan  and
      Bali, Kalika",
    booktitle = "Findings of the Association for Computational Linguistics: EMNLP 2023",
    month = dec,
    year = "2023",
    address = "Singapore",
    publisher = "Association for Computational Linguistics",
    url = "https://aclanthology.org/2023.findings-emnlp.378/",
    doi = "10.18653/v1/2023.findings-emnlp.378",
    pages = "5687--5711"
}

@misc{schulman2017proximal,
      title={Proximal Policy Optimization Algorithms}, 
      author={John Schulman and Filip Wolski and Prafulla Dhariwal and Alec Radford and Oleg Klimov},
      year={2017},
      eprint={1707.06347},
      archivePrefix={arXiv},
      primaryClass={cs.LG},
      url={https://arxiv.org/abs/1707.06347}, 
}

@article{wei2022chain,
  title={Chain-of-thought prompting elicits reasoning in large language models},
  author={Wei, Jason and Wang, Xuezhi and Schuurmans, Dale and Bosma, Maarten and Xia, Fei and Chi, Ed and Le, Quoc V and Zhou, Denny and others},
  journal={Advances in neural information processing systems},
  volume={35},
  pages={24824--24837},
  year={2022}
}

@inproceedings{trivedi2022interleaving,
    title = "Interleaving Retrieval with Chain-of-Thought Reasoning for Knowledge-Intensive Multi-Step Questions",
    author = "Trivedi, Harsh  and
      Balasubramanian, Niranjan  and
      Khot, Tushar  and
      Sabharwal, Ashish",
    editor = "Rogers, Anna  and
      Boyd-Graber, Jordan  and
      Okazaki, Naoaki",
    booktitle = "Proceedings of the 61st Annual Meeting of the Association for Computational Linguistics (Volume 1: Long Papers)",
    month = jul,
    year = "2023",
    address = "Toronto, Canada",
    publisher = "Association for Computational Linguistics",
    url = "https://aclanthology.org/2023.acl-long.557/",
    doi = "10.18653/v1/2023.acl-long.557",
    pages = "10014--10037"
}

@misc{yang2024qwen2,
      title={Qwen2.5 Technical Report}, 
      author={Qwen and : and An Yang and Baosong Yang and Beichen Zhang and Binyuan Hui and Bo Zheng and Bowen Yu and Chengyuan Li and Dayiheng Liu and Fei Huang and Haoran Wei and others},
      year={2025},
      eprint={2412.15115},
      archivePrefix={arXiv},
      primaryClass={cs.CL},
      url={https://arxiv.org/abs/2412.15115}, 
}

@misc{wang2022text,
      title={Text Embeddings by Weakly-Supervised Contrastive Pre-training}, 
      author={Liang Wang and Nan Yang and Xiaolong Huang and Binxing Jiao and Linjun Yang and Daxin Jiang and Rangan Majumder and Furu Wei},
      year={2024},
      eprint={2212.03533},
      archivePrefix={arXiv},
      primaryClass={cs.CL},
      url={https://arxiv.org/abs/2212.03533}, 
}

@misc{wang2025theoryagentstoolusedecisionmakers,
      title={Toward a Theory of Agents as Tool-Use Decision-Makers}, 
      author={Hongru Wang and Cheng Qian and Manling Li and Jiahao Qiu and Boyang Xue and Mengdi Wang and Heng Ji and Kam-Fai Wong},
      year={2025},
      eprint={2506.00886},
      archivePrefix={arXiv},
      primaryClass={cs.AI},
      url={https://arxiv.org/abs/2506.00886}, 
}

@misc{yuan2023scalingrelationshiplearningmathematical,
      title={Scaling Relationship on Learning Mathematical Reasoning with Large Language Models}, 
      author={Zheng Yuan and Hongyi Yuan and Chengpeng Li and Guanting Dong and Keming Lu and Chuanqi Tan and Chang Zhou and Jingren Zhou},
      year={2023},
      eprint={2308.01825},
      archivePrefix={arXiv},
      primaryClass={cs.CL},
      url={https://arxiv.org/abs/2308.01825}, 
}

@inproceedings{prompt2025amirhossein,
    author = {Razavi, Amirhossein and Soltangheis, Mina and Arabzadeh, Negar and Salamat, Sara and Zihayat, Morteza and Bagheri, Ebrahim},
    title = {Benchmarking Prompt Sensitivity in Large Language Models},
    year = {2025},
    isbn = {978-3-031-88713-0},
    publisher = {Springer-Verlag},
    address = {Berlin, Heidelberg},
    url = {https://doi.org/10.1007/978-3-031-88714-7_29},
    doi = {10.1007/978-3-031-88714-7_29},
    booktitle = {Advances in Information Retrieval: 47th European Conference on Information Retrieval, ECIR 2025, Lucca, Italy, April 6–10, 2025, Proceedings, Part III},
    pages = {303–313},
    numpages = {11},
    location = {Lucca, Italy}
}

@inproceedings{xie-etal-2024-ask,
    title = "Ask Again, Then Fail: Large Language Models' Vacillations in Judgment",
    author = "Xie, Qiming  and
      Wang, Zengzhi  and
      Feng, Yi  and
      Xia, Rui",
    editor = "Ku, Lun-Wei  and
      Martins, Andre  and
      Srikumar, Vivek",
    booktitle = "Proceedings of the 62nd Annual Meeting of the Association for Computational Linguistics (Volume 1: Long Papers)",
    month = aug,
    year = "2024",
    address = "Bangkok, Thailand",
    publisher = "Association for Computational Linguistics",
    url = "https://aclanthology.org/2024.acl-long.577/",
    doi = "10.18653/v1/2024.acl-long.577",
    pages = "10709--10745"
}

@misc{huang2025reinforcedinternalexternalknowledgesynergistic,
      title={Reinforced Internal-External Knowledge Synergistic Reasoning for Efficient Adaptive Search Agent}, 
      author={Ziyang Huang and Xiaowei Yuan and Yiming Ju and Jun Zhao and Kang Liu},
      year={2025},
      eprint={2505.07596},
      archivePrefix={arXiv},
      primaryClass={cs.CL},
      url={https://arxiv.org/abs/2505.07596}, 
}

@inproceedings{sheng2025verl,
author = {Sheng, Guangming and Zhang, Chi and Ye, Zilingfeng and Wu, Xibin and Zhang, Wang and Zhang, Ru and Peng, Yanghua and Lin, Haibin and Wu, Chuan},
title = {HybridFlow: A Flexible and Efficient RLHF Framework},
year = {2025},
isbn = {9798400711961},
publisher = {Association for Computing Machinery},
address = {New York, NY, USA},
url = {https://doi.org/10.1145/3689031.3696075},
doi = {10.1145/3689031.3696075},
booktitle = {Proceedings of the Twentieth European Conference on Computer Systems},
pages = {1279–1297},
numpages = {19},
location = {Rotterdam, Netherlands},
series = {EuroSys '25}
}

@inproceedings{kwon2023efficient,
  title={Efficient Memory Management for Large Language Model Serving with PagedAttention},
  author={Woosuk Kwon and Zhuohan Li and Siyuan Zhuang and Ying Sheng and Lianmin Zheng and Cody Hao Yu and Joseph E. Gonzalez and Hao Zhang and Ion Stoica},
  booktitle={Proceedings of the ACM SIGOPS 29th Symposium on Operating Systems Principles},
  year={2023}
}

@misc{cogkernal,
      title={Cognitive Kernel-Pro: A Framework for Deep Research Agents and Agent Foundation Models Training}, 
      author={Tianqing Fang and Zhisong Zhang and Xiaoyang Wang and Rui Wang and Can Qin and Yuxuan Wan and Jun-Yu Ma and Ce Zhang and Jiaqi Chen and Xiyun Li and Hongming Zhang and Haitao Mi and Dong Yu},
      year={2025},
      eprint={2508.00414},
      archivePrefix={arXiv},
      primaryClass={cs.AI},
      url={https://arxiv.org/abs/2508.00414}, 
}

@misc{webaggregator,
      title={Explore to Evolve: Scaling Evolved Aggregation Logic via Proactive Online Exploration for Deep Research Agents}, 
      author={Rui Wang and Ce Zhang and Jun-Yu Ma and Jianshu Zhang and Hongru Wang and Yi Chen and Boyang Xue and Tianqing Fang and Zhisong Zhang and Hongming Zhang and Haitao Mi and Dong Yu and Kam-Fai Wong},
      year={2025},
      eprint={2510.14438},
      archivePrefix={arXiv},
      primaryClass={cs.CL},
      url={https://arxiv.org/abs/2510.14438}, 
}

@article{li2026cso,
  author       = {Mukai Li and
                  Qingcheng Zeng and
                  Tianqing Fang and
                  Zhenwen Liang and
                  Linfeng Song and
                  Qi Liu and
                  Haitao Mi and
                  Dong Yu},
  title        = {Verified Critical Step Optimization for {LLM} Agents},
  journal      = {CoRR},
  volume       = {abs/2602.03412},
  year         = {2026},
  url          = {https://doi.org/10.48550/arXiv.2602.03412},
  doi          = {10.48550/ARXIV.2602.03412},
  eprinttype   = {arXiv},
  eprint       = {2602.03412},
  bibsource    = {dblp computer science bibliography, https://dblp.org}
}

@inproceedings{li-etal-2025-search,
    title = "Search-o1: Agentic Search-Enhanced Large Reasoning Models",
    author = "Li, Xiaoxi  and
      Dong, Guanting  and
      Jin, Jiajie  and
      Zhang, Yuyao  and
      Zhou, Yujia  and
      Zhu, Yutao  and
      Zhang, Peitian  and
      Dou, Zhicheng",
    editor = "Christodoulopoulos, Christos  and
      Chakraborty, Tanmoy  and
      Rose, Carolyn  and
      Peng, Violet",
    booktitle = "Proceedings of the 2025 Conference on Empirical Methods in Natural Language Processing",
    month = nov,
    year = "2025",
    address = "Suzhou, China",
    publisher = "Association for Computational Linguistics",
    url = "https://aclanthology.org/2025.emnlp-main.276/",
    doi = "10.18653/v1/2025.emnlp-main.276",
    pages = "5420--5438",
    ISBN = "979-8-89176-332-6"
}

\appendix

\section{Protocols}
\label{appendix:notation}

The definitions of the notations are summarized in Table \ref{table:data_stat}.

\begin{table*}[!ht]
  \centering
  {\begin{tabular}{cc}
    \toprule
    \textbf{Notation} & \textbf{Description} \\
    \midrule
    \rowcolor{platinum}
    \multicolumn{2}{c}{\textbf{{Internal/External Knowledge of Search Agent (Sec.~\ref{ssec:know_agent})}}} \\
    \midrule
    $\boldsymbol{\mathcal{M}}_{\theta}$ & LLM agent with parameter $\theta$. \\
    $\boldsymbol{\mathcal{K}}_I$ & Internal parametric knowledge of LLM. \\
    $\boldsymbol{\mathcal{K}}_E$ & External knowledge of retrieved documents. \\
    \midrule
    \rowcolor{platinum}
    \multicolumn{2}{c}{\textbf{{Markov Decision Process of Agent (Sec.~\ref{ssec:mdp_agent})}}} \\
    \midrule
    $\boldsymbol{\mathcal{S}}$ & Set of states where $i$-th state $\boldsymbol{s}_i\in \boldsymbol{\mathcal{S}}$. \\
    $\boldsymbol{\mathcal{A}}$ & Set of actions where $i$-th action $\boldsymbol{a}_i\in \boldsymbol{\mathcal{A}}$. \\
    $\boldsymbol{\mathcal{O}}$ & Set of observations where $i$-th observation $\boldsymbol{o}_i\in \boldsymbol{\mathcal{O}}$. \\
    $\boldsymbol{\mathcal{T}}$ & Set of state transition functions where $T(\boldsymbol{s}_{n+1}|\boldsymbol{s}_n,\boldsymbol{a}_n)\in \boldsymbol{\mathcal{T}}$. \\
    $\boldsymbol{\mathcal{R}}$ & Reward model where $r=\boldsymbol{\mathcal{R}}(\cdot)$. \\
    $\boldsymbol{\mathcal{E}}$ & External retriever or search engine. \\
    $\boldsymbol{x}$ & Input query or task. \\
    $\boldsymbol{s}_i$ & $i$-th state including query, and previous actions and observations. \\
    $\boldsymbol{\tau}$ & Completed trajectory to the query by LLM. \\
    $\boldsymbol{y}$ & LLM generated final answer of $\boldsymbol{\tau}$. \\
    $\boldsymbol{\hat y}$ & Ground-truth reference answer to query $\boldsymbol{x}$. \\
    \midrule
    \rowcolor{platinum}
    \multicolumn{2}{c}{\textbf{{Multi-turn Search Agent (Sec.~\ref{ssec:mt_agent})}}} \\
    \midrule
    $\boldsymbol{K}_b$ & External knowledge base. \\
    $\boldsymbol{d}_i$ & Retrieved Top-$K$ documents $[\boldsymbol{d}_{i,1},\dots,\boldsymbol{d}_{i,K}]$ at round $i$. \\
    $\boldsymbol{q}_i$ & Search query of $\boldsymbol{a}_i$ in round $i$. \\
    $\boldsymbol{p}_i$ & Thinking step of $\boldsymbol{a}_i$ in round $i$. \\
    \midrule
    \rowcolor{platinum}
    \multicolumn{2}{c}{\textbf{{Context Refiner for Search Agent (Sec.~\ref{ssec:refiner})}}} \\
    \midrule
    $\boldsymbol{\mathcal{F}}$ & Context refiner. \\
    $\boldsymbol{\tilde d}_i$ & Refined documents with critical information extracted from $\boldsymbol{d}_i$ to $\boldsymbol{q}_{i-1}$. \\
    $\boldsymbol{\mathcal{D}}_d$ & Distill dataset. \\
    $\boldsymbol{\mathcal{D}}_c$ & Context refinement dataset. \\
    $\boldsymbol{\mathcal{M}}_{\pi}$ & Foundation LLM. \\
    $\boldsymbol{\mathcal{M}}_{T}$ & Advanced teacher model. \\
    $\mathcal{L}^{\text{SFT}}_{\pi}$ & Loss function to train context refiner using SFT. \\
    \midrule
    \rowcolor{platinum}
    \multicolumn{2}{c}{\textbf{{Context Refinement for Search Agent Training (Sec.~\ref{sec:refinement})}}} \\
    \midrule
    $\mathcal{L}^{\text{CRRL}}_{\pi}$ & Loss function to train policy model based on GRPO. \\
    $\boldsymbol{\mathcal{M}}_{\pi}$ & Policy model. \\
    $\boldsymbol{\mathcal{M}}_{\text{ref}}$ & Reference model. \\
    $G$ & Number of rollouts in one group. \\
    $\epsilon$ & Clipping ratio. \\
    $\beta$ & Coefficient for the KL divergence. \\
    $A_j$ & Advantage estimate based on the group-relative rewards of the trajectory. \\
    $\boldsymbol{\mathcal{D}}_t$ & Training dataset for GRPO. \\
    \bottomrule
  \end{tabular}}
  \caption{Summarized notations in this work.}
\label{table:notation}
\end{table*}

\section{Related Work}
\label{append:related}

\subsection{LLM-based Search Agent}
\label{ssec:search_agent}
Although LLMs exhibit impressive capabilities, they often lack updated or domain-specific knowledge \citep{peng2023study,li2023large}, which undermines LLMs' reliability.
Therefore, search engines \citep{zhao2024dense} are widely integrated to provide external evidence.
A common paradigm is retrieval-augmented generation (RAG) \citep{gao2024retrievalaugmentedgenerationlargelanguage,lewis2020retrieval}, in which a search engine retrieves documents relevant to the search query and feeds them into LLMs.
More recent work treats search engines as interactive tools \citep{schick2023toolformer}, prompting or fine-tuning LLMs to act as search agents \citep{xiong2025rag,cogkernal,webaggregator,li2026cso}.
Approaches such as IRCoT \citep{trivedi2023interleaving} and ReAct \citep{yaoreact} employ prompting to interleave reasoning with iterative search calls.
Search-R1 \citep{jin2025search} optimizes LLMs to produce high-quality trajectories through real-time, multi-turn search interactions using RL.
Despite these advances, prior studies have largely focused on eliciting LLMs to output accurate knowledge in reasoning paths despite given lengthy contexts with noise, while overlooking the contextual interference introduced by multi-turn search interactions, which may cascade across subsequent actions, degrading the reliability and efficiency of the final answers for search agents.

\subsection{Contextual Interference}
\label{ssec:context_interference}
LLMs are highly sensitive to input contexts, leaving them vulnerable to noise or irrelevant content that can degrade output quality \citep{xie-etal-2024-ask,prompt2025amirhossein,webaggregator}.
Existing strategies to mitigate this issue fall into three main categories.
\textbf{(1) Key Information Extraction} identifies and preserves the most critical or relevant information in contents as in dialogue state tracking \citep{jacqmin-etal-2022-follow} or RAG reranking \citep{glass-etal-2022-re2g,nguyen2025maragmultiagentretrievalaugmentedgeneration,yu2024rankragunifyingcontextranking}; 
\textbf{(2) Compression method} condenses lengthy input sequences into summaries or latent states to weaken noise \citep{jiang2023llmlingua,yi2025surveyrecentadvancesllmbased,li2025singleturnsurveymultiturninteractions}. 
\textbf{(3) Prompt-based method} explicitly instructs LLMs to disregard irrelevant content \citep{rajeev2025catsconfusereasoningllm}.
However, Key Information Extraction relies on task-specific schema, Compression may distort key information and incur extra compression cost, and Prompt-based methods are sensitive to prompt design.
More importantly, these issues are exacerbated in multi-turn search-agent scenarios, where the large volume of retrieved documents after several turns introduces more noise and irrelevant information.

\subsection{Reinforcement Learning for Agent}
\label{ssec:rl_agent}
Reinforcement Learning (RL) \citep{kaelbling1996reinforcement} has emerged as a paradigm for LLM post-training or alignment \citep{ouyang2022training}.
A variety of RL algorithms have been introduced, like PPO \citep{schulman2017proximal} and GRPO \citep{shao2024deepseekmath}. 
With specific environments and reward designs, LLMs can evolve as autonomous agents capable of adaptive decision-making and interactions with the environment.
One representative application is Search Agent \citep{song2025r1searcherincentivizingsearchcapability,jin2025search,chen2025learning}, which interacts with search engines to iteratively gather external knowledge into its reasoning, and thereby performs knowledge-intensive tasks more effectively.
However, current RL research focuses on deriving the optimal action from rollout trajectories while overlooking the potential influence of context interference within trajectories—particularly in search agents—thereby constraining the achievable performance of RL algorithms.

\section{Dataset Details}
\label{appendix:dataset}

Experiments are conducted to evaluate the performance of search agents on various closed-book QA datasets, which necessitate extra retrieval to address, encompassing both single- and multi-hop scenarios.
\textbf{Single-hop QA} includes:
1) \textbf{Natural Questions (NQ)} \citep{kwiatkowski2019natural}, which is constructed by Google Search queries along with annotated short answers;
2) \textbf{TriviaQA} \citep{joshi2017triviaqa}, which contains closed-book trivia QA pairs to gauge models’
factual knowledge; 
and 3) \textbf{PopQA} \citep{mallen2022not}, which consists of entity-centric QA pairs converted from a knowledge tuple retrieved in Wikidata.
\textbf{Multi-hop QA} includes:
1) \textbf{HotpotQA} \citep{yang2018hotpotqa}, the first large-scale dataset requiring reasoning across multiple Wikipedia paragraphs; 
2) \textbf{2WikiMultiHopQA (2Wiki)} \citep{ho2020constructing}, which provides evidence information containing a reasoning path for multi-hop questions; 
3) \textbf{MuSiQue} \citep{trivedi2022musique}, which features more difficult 2-4 hop questions; 
and 4) \textbf{Bamboogle} \citep{press2022measuring}, which is made up only of complex questions that Google answers incorrectly.
Dataset statistics of seven test sets are presented in Table \ref{table:data_stat}.

\begin{table*}[!t]
  \centering
  {\begin{tabular}{ccccccccc}
    \toprule
    \textbf{Dataset} & \textbf{NQ} & \textbf{TriviaQA} & \textbf{PopQA} & \textbf{HotpotQA} & \textbf{2Wiki} & \textbf{MiSique} & \textbf{Bamboogle} \\
    \midrule
    \bf \# Ques. & 3610 & 11313 & 14267 & 7405 & 12576 & 2417 & 125 \\
    \bottomrule
  \end{tabular}}
  \caption{Data statistics of questions in seven test sets.}
\label{table:data_stat}
\end{table*}

\begin{figure*}[!t]
    \centering
    \includegraphics[width=0.75\linewidth]{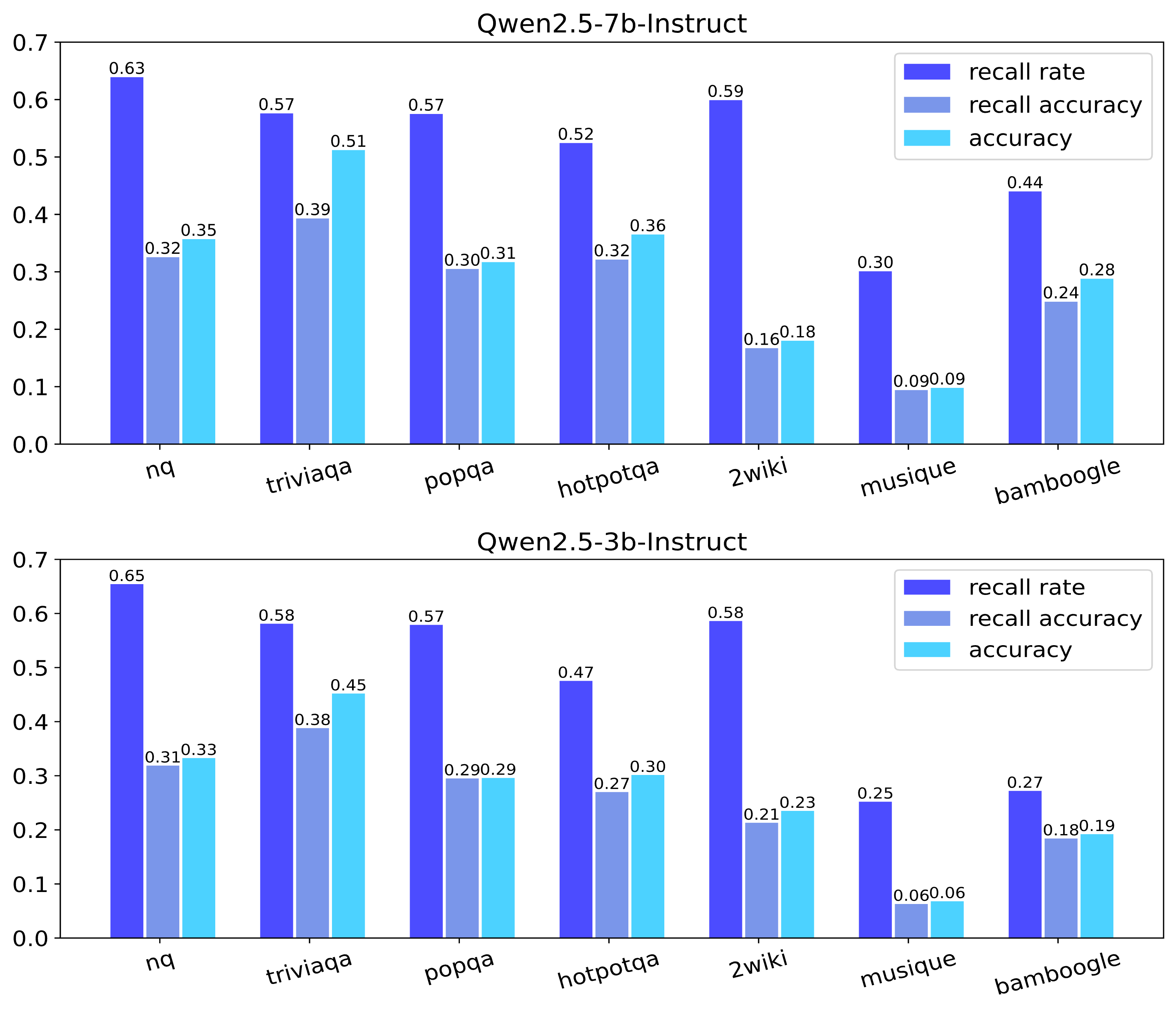}
    \caption{Demonstration of how contextual interference affects the performance of LLM search agents. ``recall rate''=$N_\text{r}/N$ denotes the proportion of questions for which the retrieved documents contain the correct answer ($N_\text{r}$) among all questions ($N$). ``recall accuracy''=$N_\text{rc}/N$ refers to the proportion of correctly answered questions ($N_\text{rc}$) in $N_\text{r}$, relative to all questions ($N$). ``accuracy'' represents the proportion of all correctly answered questions $N_\text{c}$ out of the questions ($N$).}
    \label{fig:recall_com}
\end{figure*}

\subsection{Concerns about Data Contamination}

We have carefully considered the concern about Data contamination during the experiments and verified that the external knowledge base $\boldsymbol{K}_b$ will not contaminate the internal knowledge $\boldsymbol{K}_I$ as follows.

In our experiments, the timeline of test sets is synchronized with the wiki dump, so $\boldsymbol{K}_b$ is regarded as the ground-truth knowledge.
Although there may be partial knowledge conflicts between $\boldsymbol{K}_b$ and $\boldsymbol{K}_I$ due to knowledge updates, questions in test sets usually have clear timeline information as presented below.

\begin{quote}
\texttt{\textbf{HotpotQA}: The 1895/96 Football League season was the eighth in Football League history with Everton, their Goodison Park home, is a football stadium located in Walton, Liverpool, in which country?}
\end{quote}

\begin{quote}
\texttt{\textbf{PopQA}: In the 80s who wrote the novel Empire of The Sun?}
\end{quote}

Therefore, $\boldsymbol{K}_b$ will serve as a ground truth and will not contaminate $\boldsymbol{K}_I$.
A very small number of ambiguous questions can be ignored and will not affect the performance evaluation.
The search agent framework and evaluation setting of this work are implemented based on Search-R1, which is reliable and widely adopted by a series of studies.

\section{Training Details}
\label{appendix:implementation}

For GRPO training, we set the policy LLM learning rate to 1e-6 and sample 4 responses per prompt, following the GRPO implementation in Verl \citep{sheng2025verl}.
The batch size is set at 32, with a mini-batch size of 8 and a micro-batch size of 4.
The maximum input sequence length and generation length are set to 2048 and 500 respectively.
We enable gradient checkpointing and use Fully Sharded Data Parallel (FSDP) with CPU offloading.
For efficient LLM rollouts, we adopt vLLM \citep{kwon2023efficient} with a tensor parallel size of 1 and GPU memory utilization ratio of 0.7.
The rollout sampling temperature is set to 0.7 and the top-p value to 1.0.
The KL divergence regularization coefficient $\beta$ and clip ratio $\epsilon$ are set to 0.001 and 0.2.
The maximum action budget $B$ is set to 8.
In cases where training diverges, we evaluate at the most recent stable checkpoint according to the training reward curve; otherwise, the final checkpoint is used for evaluation.

For training-based baselines, due to the computational resource limitation with only 4×40G A100 GPU cards, fine-tuning on the all corpus of 160k samples of the training corpus is expensive. 
Therefore, we employ a subset with 60k samples randomly sampled from the original training corpus. 
All other training settings are maintained. 
Experiments conducted using CRRL are to validate the effectiveness of incorporating context refinement in RL pipelines for search agent training, which is regardless of the training corpus quantity. 
We will further add and clarify details of the training setting differences in the final version of the manuscript.

\section{Experiments}
\label{append:experiments}

\subsection{Context Interference Effects}

We have presented the results of context interference effects on “recall rate” and “recall accuracy” of context refinement and baselines in Table \ref{table:qwen-recall} as follows.

\begin{table}[!ht]
    \centering
    \footnotesize
    \resizebox{.5\textwidth}{!}
    {
    \begin{tabular}{lcccccccccccc}
        \toprule
         & {\textbf{recall rate}} & {\textbf{recall accuracy}} & {\textbf{accuracy}} \\
        \hline
        \rowcolor{seashell}
        \multicolumn{9}{c}{\textbf{{Qwen2.5-3b-Instruct}}} \\
        \hline
        \bf IRCoT & 48.5 & 22.3 & 24.3 \\
        \bf GPT-Compress & 48.6 & 24.7 & 26.4 \\
        \bf GPT-Refine & 46.9 & 26.9 & 28.7 \\
        \bf Self-Refine & 46.5 & 22.4 & 24.5 \\
        \bf Context Refiner & 46.7 & 24.3 & 26.6 \\
        \hline
        \bf RFT & 49.1 & 24.7 & 26.7 \\
        \bf Search-GRPO & 49.4 & 28.1 & 29.8 \\
        \bf CRRL & 48.4 & 29.7 & 31.5 \\
        \hline
        \rowcolor{seashell}
        \multicolumn{9}{c}{\textbf{{Qwen2.5-7b-Instruct}}} \\
        \hline
        \bf IRCoT & 52.6 & 24.4 & 27.5 \\
        \bf GPT-Compress & 51.5 & 27.2 & 30.5 \\
        \bf GPT-Refine & 50.0 & 29.4 & 33.2 \\
        \bf Self-Refine & 50.6 & 26.3 & 29.2 \\
        \bf Context Refiner & 51.2 & 28.6 & 32.2 \\
        \hline
        \bf IRCoT & 52.6 & 24.4 & 27.5 \\
        \bf RFT & 53.5 & 28.8 & 33.1 \\
        \bf Search-GRPO & 54.9 & 32.2 & 34.6 \\
        \bf CRRL & 54.4 & 34.3 & 36.6 \\
        \bottomrule
    \end{tabular}}
    \caption{Context interference effects on ``recall rate'' and ``recall accuracy'' of context refinement and baseline methods, measured by Exact Match (EM).}
    \label{table:qwen-recall}
\end{table}

\subsection{Alternative Ranking Baselines}

\paragraph{Baseline Setting}
To avoid the interference derived from the irrelevant/distracting search results, we employ alternative ranking baselines.
In our experiments, the retriever will consistently return the top-3 highest-scoring documents, which may contain irrelevant documents with relatively lower retrieval scores.
The retrieval score ranges from [0, 1]. Therefore, we set different thresholds for retrieval scores to filter out irrelevant documents. 
We report the performance (EM/ART) of the ranking baseline with other context refinement methods in Table \ref{table:ranking} for interference mitigation as follows.
EM and ART denote the Exact Match (EM) and Average Retrieval Times (ART), respectively.

\begin{table*}[!t]
    \centering
    \footnotesize
    {
    \begin{tabular}{lcccccccccccc}
        \toprule
        \multirow{2}{*}{\textbf{Methods}} & \multicolumn{3}{c}{\textbf{Single-Hop QA}} & \multicolumn{4}{c}{\textbf{Multi-Hop QA}} & \multirow{2}{*}{\textit{\textbf{Avg.}}} \\
        \cmidrule(r){2-4}\cmidrule(r){5-8}
         & \textbf{NQ} & \textbf{TriviaQA} & \textbf{PopQA} & \textbf{HotpotQA} & \textbf{2wiki} & \textbf{Musique} & \textbf{Bamboogle} & \\
        \hline
        \hline
        \rowcolor{seashell}
        \multicolumn{9}{c}{\textbf{{Qwen2.5-7b-Instruct}}} \\
        \hline
        \bf IRCoT & 30.6~/~2.0 & 51.2~/~1.8 & 31.7~/~2.1 & 24.6~/~3.0 & 18.0~/~3.4 & 9.8~/~3.1 & 28.8~/~2.5 & 27.5~/~2.6 \\
        \hdashline
        \bf GPT-Compress & 32.5~/~\textbf{0.9} & 55.5~/~1.0 & 34.8~/~1.1 & 29.8~/~1.6 & 25.0~/~1.5 & 9.6~/~1.3 & 26.4~/~1.1 & 30.5~/~\textbf{1.2} \\
        \bf GPT-Refine & \textbf{34.6}~/~1.1 & \textbf{57.5}~/~\textbf{0.9} & \textbf{37.0}~/~1.0 & \textbf{33.0}~/~\textbf{1.3} & \textbf{27.6}~/~1.8 & 9.3~/~1.4 & \textbf{33.6}~/~1.0 & \textbf{33.2}~/~1.2 \\
        \bf Self-Refine & 32.5~/~1.5 & 53.3~/~0.9 & 33.0~/~\textbf{1.0} & 27.0~/~1.4 & 21.6~/~\textbf{1.5} & 8.4~/~\textbf{1.2} & 28.8~/~\textbf{1.0} & 29.2~/~1.2 \\
        \hdashline
        \bf Context Refiner & 34.0~/~1.2 & 56.9~/~0.9 & 36.4~/~1.0 & 32.0~/~1.4 & 27.2~/~1.6 & 8.6~/~1.2 & 30.4~/~1.0 & 32.2~/~1.2 \\
        \hdashline
        \bf Ranking (t=0.2) & 30.8~/~2.0 & 51.2~/~1.8 & 31.9~/~2.2 & 24.6~/~3.2 & 18.2~/~3.3 & 9.8~/~3.0 & 28.8~/~2.6 & 27.9~/~2.6 \\
        \bf Ranking (t=0.5) & 31.2~/~1.8 & 51.4~/~2.0 & 31.8~/~2.4 & 25.0~/~3.3 & 18.3~/~3.5 & 9.8~/~3.2 & 28.8~/~2.6 & 28.0~/~2.7 \\
        \hline
        \rowcolor{seashell}
        \multicolumn{9}{c}{\textbf{{Qwen2.5-3b-Instruct}}} \\
        \hline
        \bf IRCoT & 21.6~/~1.2 & 45.2~/~1.1 & 29.6~/~1.1 & 24.1~/~1.6 & 23.5~/~1.9 & 6.8~/~1.6 & 19.2~/~1.4 & 24.3~/~1.4 \\
        \hdashline
        \bf GPT-Compress & 31.3~/~\textbf{0.9} & 51.5~/~0.9 & 29.9~/~1.0 & 24.1~/~1.1 & 22.7~/~\textbf{1.2} & 7.2~/~1.0 & 18.4~/~0.9 & 26.4~/~1.0 \\
        \bf GPT-Refine & \textbf{32.6}~/~1.0 & \textbf{54.5}~/~\textbf{0.8} & \textbf{35.6}~/~\textbf{0.9} & \textbf{24.4}~/~\textbf{1.0} & 22.5~/~1.3 & \textbf{7.5}~/~\textbf{0.9} & \textbf{24.0}~/~\textbf{0.9} & \textbf{28.7}~/~\textbf{1.0} \\
        \bf Self-Refine & 23.7~/~1.1 & 47.6~/~0.9 & 30.8~/~0.9 & 23.1~/~1.0 & 21.4~/~1.2 & 5.0~/~1.1 & 20.0~/~0.9 & 24.5~/~1.0 \\
        \hdashline
        \bf Context Refiner & 30.7~/~1.0 & 50.8~/~0.9 & 32.0~/~0.9 & 24.1~/~1.0 & 22.2~/~1.2 & 7.2~/~1.0 & 19.4~/~0.9 & 26.6~/~1.0 \\
        \hdashline
        \bf Ranking (t=0.2) & 21.7~/~1.2 & 45.4~/~1.1 & 29.8~/~1.1 & 24.3~/~1.7 & 23.7~/~1.9 & 6.8~/~1.6 & 19.2~/~1.4 & 24.4~/~1.4 \\
        \bf Ranking (t=0.5) & 21.9~/~1.3 & 45.8~/~1.2 & 30.2~/~1.3 & 24.4~/~1.8 & 23.6~/~1.9 & 6.8~/~1.7 & 19.2~/~1.4 & 24.6~/~1.5 \\
        \hline
        \bottomrule
    \end{tabular}}
    \caption{Performance results of EM/ART across QA test sets on several baselines as well as our proposed CRRL method, measured by Exact Match (EM) and Average Retrieval Times (ART).}
    \label{table:ranking}
\end{table*}

\paragraph{Analysis}
The ranking baselines marginally outperform the IRCoT but underperform other context refinement methods, which can be attributed that

1. Ranking baselines can effectively filter out irrelevant documents to mitigate context interference but can also remove the ground-truth documents. Therefore, they can not lead to consistent performance improvements in both reliability and efficiency.

2. The performance of ranking baselines on different cases rely on the threshold setting, which lacks flexibility compared with other model-based context refinement methods.

\section{Prompt Template}
\label{append:prompt}

\begin{quote}
    \begin{tcolorbox}[colback=almond!20!white, colframe=almond!60!black, title=\textbf{Direct Inference Prompt}]
    \label{temp:icl}
        You are an excellent Question-Answering assistant. Please answer the following question based on your knowledge. You can directly provide the answer inside \answer{and}, without detailed illustrations. For example, \answer{North America}. Question: \{\texttt{question}\}\\
    \end{tcolorbox}
    
    \begin{tcolorbox}[colback=almond!20!white, colframe=almond!60!black, title=\textbf{CoT Prompt}]
    \label{temp:icl-idk}
        You are an excellent Question-Answering assistant. Please answer the following question based on your knowledge. You must conduct reasoning inside \think{and} to think step by step first. You can directly provide the answer inside \answer{and}, without detailed illustrations. For example, \answer{North America}. Question: \{\texttt{question}\}\\
    \end{tcolorbox}
    
    \begin{tcolorbox}[colback=almond!20!white, colframe=almond!60!black, title=\textbf{IRCoT Prompt}]
    \label{temp:icl-cot}
        You are an excellent Question-Answering assistant. Please answer the following question based on your knowledge. You must conduct reasoning inside \think{and} to think step by step first. After reasoning, if you find you lack some knowledge, you can call a search engine by \search{and} and it will return the top searched results between \info{and}. You can search as many times as your want. If you find no further external knowledge needed, you can directly provide the answer inside <answer> and </answer>, without detailed illustrations. For example, \answer{North America}. Question: \{\texttt{question}\}\\
    \end{tcolorbox}
\end{quote}

\begin{table*}[ht]
    \centering
    \footnotesize
    \begin{tabular}{p{15.6cm}}
        \toprule
        \textbf{Question} $\boldsymbol{x}$: Who is the mother of the father of George Washington? \\
        \midrule
        \rowcolor{pearl}
        \textbf{ID 1: Generation using full documents with contextual interference.} \\
        \hdashline
        \textbf{Generation Round 1} $[\boldsymbol{a}_0, \boldsymbol{o}_1]$: \think{To find out who the mother of the father of George Washington is, I need to trace back the family lineage of George Washington.}
        \search{Who is the mother of the father of George Washington?}\\
        \info{\textbf{Doc 1}(Title: ``Lawrence Washington (1659–1698)'') Lawrence Washington (1659–1698) Lawrence Washington (1659 – February 1698), a colonial-era American who is principally remembered as the paternal grandfather of George Washington. He was the owner of a substantial Virginia plantation that he inherited from his father, John Washington, as the firstborn son under the law of primogeniture. Washington was sent to England to finish his education. In addition to being a landowner and planter, he was a lawyer, soldier, and a politician in colonial Virginia. Lawrence Washington was the eldest son of Lt. Col. John Washington (from Essex, England) and Anne Pope. He was born in 1659, on ... \textbf{Doc 2}(Title: ``Augustine Washington'') Augustine Washington Augustine Washington Sr. (November 12, 1694 – April 12, 1743) was the father of the first U.S. President George Washington. He belonged to the Colony of Virginia's landed gentry and was a planter and slaveholder. Augustine Washington was born in Westmoreland County, Virginia, on November 12, 1694 to Capt. Lawrence Washington, a militia captain and a member of the Virginia House of Burgesses, and Mildred Warner. His paternal grandparents were Lt. Col. John Washington (c. 1631–1677) and his first wife, Anne Pope. Augustine was only four years old when his father died. He inherited about on Bridges Creek ... \textbf{Doc 3}(Title: ``Lawrence Washington (1602–1653)'') Lawrence Washington (1602–1653) Rev. Lawrence Washington (1602 – 21 January 1653) was an English rector, and the great-great-grandfather of George Washington. Washington was born in 1602. He was the fifth son of Lawrence Washington (1565–1616) of Sulgrave Manor, Northamptonshire, son and heir of Robert Washington (1544–1619) esquire, of Sulgrave by his first wife Elizabeth Lyte, daughter and heiress of Walter Lyte of Radway, Warwickshire. His mother was Margaret Butler (d. 16 March 1651), the eldest daughter and co-heiress of William Butler, esquire, of Tyes Hall in Cuckfield, Sussex, and Margaret Greeke, the daughter of Thomas Greeke, gentleman, of Palsters, Lancashire.} \\ \\
        \textbf{Generation Round 2} $[\boldsymbol{a}_1]$: \answer{Anne Popej} \\
        \midrule
        \rowcolor{platinum}
        \textbf{ID 2: Generation using our proposed.} \\
        \hdashline
        \textbf{Generation Round 1} $[\boldsymbol{a}_0, \boldsymbol{o}_1]$: \think{To solve this problem, I need to trace back the lineage of George Washington's family. George Washington was the first President of the United States. His father was named Augustine Washington. So, my next step would be to identify who Augustine Washington's mother was.} \search{Who was the mother of George Washington's father Augustine Washington?} \\
        \info{Augustine Washington, the father of George Washington, was born to Capt. Lawrence Washington and Mildred Warner. Therefore, the mother of Augustine Washington was Mildred Warner.} \\ \\
        \textbf{Generation Round 2} $[\boldsymbol{a}_1]$: 
        \answer{Mildred Warner} \\
\bottomrule
\end{tabular}
    \caption{Demonstrations of one generation of a search agent given original retrieved documents (\textbf{ID 1}) and extracted information (\textbf{ID 2}) respectively.}\label{tab:case2}
\end{table*}

\begin{table*}[ht]
    \centering
    \footnotesize
    \begin{tabular}{p{15.6cm}}
        \toprule
        \textbf{Question} $\boldsymbol{x}$: Hndrxx features a guest appearance by what Canadian singer? \\
        \midrule
        \rowcolor{pearl}
        \textbf{ID 1: Generation using full documents with contextual interference.} \\
        \hdashline
        \textbf{Generation Round 1} $[\boldsymbol{a}_0, \boldsymbol{o}_1]$: \think{I need to find out which Canadian singer Hndrxx has featured in his song.}
        \search{Hndrxx featuring Canadian singer}\\
        \info{\textbf{Doc 1}(Title: Hndrxx) Hndrxx Hndrxx (stylized in all caps and pronounced ``Hendrix'') is the sixth studio album by American rapper Future. It was released on February 24, 2017, by A1 Recordings, Epic Records, and Freebandz. It followed the release of Future's fifth eponymously titled album, ``Future'' (2017), by one week, and features guest appearances from Rihanna, The Weeknd, Chris Brown, and Nicki Minaj. ``Hndrxx'' is also eponymously named after Future's alter ego, Future Hendrix. ``Hndrxx'' was supported by four singles: ``Selfish'', ``Pie'', ``Incredible'', and ``You da Baddest''. The album received positive critical reviews and charted at number one, making Future the first artist ... \textbf{Doc 2}(Title: Hndrxx) featuring Trinidadian-American rapper Nicki Minaj, was released as the album's fourth single on July 28, 2017. The song was produced by Go Grizzly and Detail. ``Hndrxx'' received generally positive reviews from critics. At Metacritic, which assigns a normalized rating out of 100 to reviews from mainstream publications, the album received an average score of 78, based on eight reviews. Neil Z. Yeung of AllMusic said, ``Released'' just a week after his self-titled fifth effort, Future's ``HNDRXX'' provided an introspective and confessional complement to the more extroverted ``Future''. Ural Garrett of ``HipHopDX'' said, ``HNDRXX'' provides a view into a modern rock ... \textbf{Doc 3}(Title: Hndrxx) included on to the tracklist on streaming services. On July 28, 2017, another song was added to the streaming tracklist, ``You da Baddest''. ``Selfish'' featuring Barbadian singer Rihanna, was released as the album's lead single on February 28, 2017. The song was produced by Detail, Major Seven and Mantra. ``Pie'' featuring American singer Chris Brown, was released as the second single on June 25, 2017. The song was produced by Detail and D. A. Doman. The album's third single, ``Incredible'', was released to rhythmic contemporary radio on July 25, 2017. The song was produced by Dre Moon. ``You da Baddest''} \\ \\
        \textbf{Generation Round 2} $[\boldsymbol{a}_1, \boldsymbol{o}_2]$: \think{From the information provided, Hndrxx features a guest appearance by Rihanna, Chris Brown, and Nicki Minaj. However, the specific Canadian singer who appears in Hndrxx is not mentioned here. I need to search again for the exact information.} \search{Hndrxx Canadian singer guest appearance} \\
        \info{\textbf{Doc 1}(Title: Hndrxx) Hndrxx Hndrxx (stylized in all caps and pronounced ``Hendrix'') is the sixth studio album by American rapper Future. It was released on February 24, 2017, by A1 Recordings, Epic Records, and Freebandz. It followed the release of Future's fifth eponymously titled album, ``Future'' (2017), by one week, and features guest appearances from Rihanna, The Weeknd, Chris Brown, and Nicki Minaj. ``Hndrxx'' is also eponymously named after Future's alter ego, Future Hendrix. ``Hndrxx'' was supported by four singles: ``Selfish'', ``Pie'', ``Incredible'', and ``You da Baddest''. The album received positive critical reviews and charted at number one, making Future the first artist ... \textbf{Doc 2}(Title: Hndrxx) included on to the tracklist on streaming services. On July 28, 2017, another song was added to the streaming tracklist, ``You da Baddest''. ``Selfish'' featuring Barbadian singer Rihanna, was released as the album's lead single on February 28, 2017. The song was produced by Detail, Major Seven and Mantra. ``Pie'' featuring American singer Chris Brown, was released as the second single on June 25, 2017. The song was produced by Detail and D. A. Doman. The album's third single, ``Incredible'', was released to rhythmic contemporary radio on July 25, 2017. The song was produced by Dre Moon. ``You da Baddest'' ... \textbf{Doc 3}(Title: Hndrxx) featuring Trinidadian-American rapper Nicki Minaj, was released as the album's fourth single on July 28, 2017. The song was produced by Go Grizzly and Detail. ``Hndrxx'' received generally positive reviews from critics. At Metacritic, which assigns a normalized rating out of 100 to reviews from mainstream publications, the album received an average score of 78, based on eight reviews. Neil Z. Yeung of AllMusic said, Released just a week after his self-titled fifth effort, Future's ``HNDRXX'' provided an introspective and confessional complement to the more extroverted ``Future''. Ural Garrett of ``HipHopDX'' said, ``HNDRXX'' provides a view into a modern rock} \\ \\
        \textbf{Generation Round 3} $[\boldsymbol{a}_2]$: \think{It seems that the search did not yield the specific information I was looking for. However, I remember from the initial searches that Hndrxx featured Rihanna, Chris Brown, and Nicki Minaj. Among these, Nicki Minaj is known to be a Canadian-American rapper. Therefore, the Canadian singer who appears on Hndrxx is likely Nicki Minaj.} \answer{Nicki Minaj} \\
        \midrule
        \rowcolor{platinum}
        \textbf{ID 2: Generation using our proposed.} \\
        \hdashline
        \textbf{Generation Round 1} $[\boldsymbol{a}_0, \boldsymbol{o}_1]$: \think{I need to know who Hndrxx is and which Canadian singer he featured with. I don't have this information, so I should search for it.} \search{Hndrxx Canadian singer guest appearance} \\
        \info{The album \"Hndrxx\" by Future features guest appearances from Rihanna, The Weeknd, Chris Brown, and Nicki Minaj. Chris Brown and Nicki Minaj are American singers. The Weeknd, a Canadian singer, is one of the guest artists on the album.} \\ \\
        \textbf{Generation Round 2} $[\boldsymbol{a}_1]$: 
        \think{I found that the Canadian singer The Weeknd is a guest artist on Hndrxx's album.} \answer{The Weeknd} \\
\bottomrule
\end{tabular}
    \caption{Demonstrations of one generation of a search agent given original retrieved documents (\textbf{ID 1}) and extracted information (\textbf{ID 2}) respectively.}\label{tab:case}
\end{table*}

\end{document}